\documentclass{article}

\usepackage{arxiv}

\usepackage[utf8]{inputenc} 
\usepackage[T1]{fontenc}    
\usepackage{url}            
\usepackage{booktabs}       
\usepackage{amsfonts}       
\usepackage{nicefrac}       
\usepackage{microtype}      
\usepackage{lipsum}		
\usepackage{graphicx}
\usepackage[numbers]{natbib}
\usepackage{hyperref} 
\hypersetup{colorlinks = true, citecolor = blue}
\usepackage{doi}
\usepackage{makecell, longtable}
\usepackage{multicol}
\usepackage[graphicx]{realboxes}
\usepackage{enumitem}
\usepackage{amsmath}
\usepackage{amsthm}
\PassOptionsToPackage{table}{xcolor}
\usepackage{tikz,ifthen,fp,calc}
\usepackage{tikzpeople}
\usetikzlibrary{shapes}
\usepackage{fontawesome} 
\usepackage{pifont} 
\usepackage{wasysym} 
\usepackage[capitalize, noabbrev, nameinlink]{cleveref}
\usepackage{colortbl} 
\usepackage{pgfplots}
\pgfplotsset{compat=1.10}
\usepgfplotslibrary{fillbetween}
\usetikzlibrary{matrix}
\usetikzlibrary{patterns}
\usetikzlibrary{backgrounds}
\usepackage{subcaption}
\usepackage{multirow}
\usepackage{tcolorbox}

\newcommand*{\R}{\mathbb{R}}

\pgfplotsset{scaled y ticks=false}

\newcommand{\PreserveBackslash}[1]{\let\temp=\\#1\let\\=\temp}
\newcolumntype{C}[1]{>{\PreserveBackslash\centering}p{#1}}
\newcolumntype{R}[1]{>{\PreserveBackslash\raggedleft}p{#1}}
\newcolumntype{L}[1]{>{\PreserveBackslash\raggedright}p{#1}}

\title{Advances in Continual Graph Learning for Anti‐Money Laundering Systems: A Comprehensive Review}

\author{
Bruno Deprez\thanks{Correspondening author: \href{mailto:bruno.deprez@kuleuven.be}{\texttt{bruno.deprez@kuleuven.be}}} \\
KU Leuven\\
University of Antwerp - imec
\And
Wei Wei\\
University of Antwerp - imec, IDLab
\And
Wouter Verbeke\\
KU Leuven 
\And
Bart Baesens\\
KU Leuven \\
University of Southampton 
\And
Kevin Mets \\
University of Antwerp - imec, IDLab
\And
Tim Verdonck\\
University of Antwerp - imec\\
KU Leuven \\
}

\begin{document}

\maketitle

\begin{abstract}
Financial institutions are required by regulation to report suspicious financial transactions related to money laundering. Therefore, they need to constantly monitor vast amounts of incoming and outgoing transactions. A particular challenge in detecting money laundering is that money launderers continuously adapt their tactics to evade detection. Hence, detection methods need constant fine-tuning. Traditional machine learning models suffer from catastrophic forgetting when fine-tuning the model on new data, thereby limiting their effectiveness in dynamic environments. Continual learning methods may address this issue and enhance current anti‐money laundering (AML) practices, by allowing models to incorporate new information while retaining prior knowledge. Research on continual graph learning for AML, however, is still scarce. In this review, we critically evaluate state‐of‐the‐art continual graph learning approaches for AML applications. We categorise methods into replay‐based, regularization‐based, and architecture-based strategies within the graph neural network (GNN) framework, and we provide in-depth experimental evaluations on both synthetic and real-world AML data sets that showcase the effect of the different hyperparameters. Our analysis demonstrates that continual learning improves model adaptability and robustness in the face of extreme class imbalances and evolving fraud patterns. Finally, we outline key challenges and propose directions for future research.
\end{abstract}

\keywords{Continual Learning \and Anti‐Money Laundering \and Graph Neural Networks \and Fraud Detection \and Catastrophic Forgetting} 

\section{Introduction}
Criminal enterprise activities generate income streams that cannot be directly used because of their illegal origins. Therefore, criminals \textit{launder money} to make illicitly obtained funds appear legitimate~\citep{levi2006money}. The \citet{UNODC} has estimated that an amount equal to about $2\%$ to $5\%$ of global GDP is laundered each year, amounting to USD~2~trillion. This money is used to expand criminal activity and to finance terrorism~\citep{levi2006money}, resulting in enormous socioeconomic pressure. 

Generally, money laundering approaches involve three main steps~\citep{UNODC, levi2006money}. During \textbf{placement}, the illegal money enters the financial system, often in jurisdictions where regulation and enforcement are less strict. \textbf{Layering} involves mixing the illegal funds with legitimately obtained money across multiple transactions. This obscures the initial source of the money, making it harder to uncover the illegal origin. At \textbf{integration}, the money is spent on legitimate purchases. After completing this final step, the money is successfully laundered. Actual money laundering approaches may also include fewer or more steps.

The framework given above illustrates that money laundering involves many payments over multiple accounts, warranting the use of network analytics~\citep{https://doi.org/10.1002/widm.1208}.  
A popular way of adopting network analytics is via graph neural networks (GNNs). GNNs have shown promising results for fighting financial crime~\citep{MOTIE2024122156} with adoption in credit card fraud detection~\citep{10.1007/978-3-030-37720-5_3,VANBELLE2022116463,van2024SHINE}, insurance fraud detection~\citep{oskarsdottir2022social, deprez2024insurance}, and anti-money laundering~\citep{deprez2024aml}. GNNs are able to detect fraudulent behaviour by learning complex, non-linear patterns in network data~\citep{MOTIE2024122156}. 

Financial networks and fraud characteristics evolve over time, necessitating the fine-tuning of these GNNs when new data comes in. This fine-tuning can cause the model to suffer from catastrophic forgetting~\citep{catastrophicForgetting,carta2021catastrophicforgettingdeepgraph,de2021continual}, where learning sequentially on \textit{new} data while discarding \textit{old} data leads to significant performance loss on earlier observations. 

Continual learning, also known as incremental learning or lifelong learning, aims to mitigate the problem of catastrophic forgetting~\citep{goodfellow2015empiricalinvestigationcatastrophicforgetting}. Research in this field typically aims to adapt regular deep learning methods or develop new methods that are able to accumulate and consolidate knowledge. One of the key assumptions that underlie continual learning, is that data is no longer (fully) accessible after models have been trained. 

Continual learning is key for effective and dynamic anti-money laundering~(AML). Financial institutions often face enormous transaction volumes that require continuous monitoring. However, they face computational constraints when implementing AML methods in practice. First, there is limited computing power and budget to update AML models, making periodical retraining from scratch impractical. Second, there are regulatory constraints on how much data can be stored and for how long.  Finally, money laundering methods, as for other types of fraud, are evolving constantly~\citep{baesens2015fraud, VANVLASSELAER201538}, so the distribution of illegitimate transactions changes over time. However, when training to detect these new tactics, models should be able to retain knowledge about old ones, in case these are used again. Otherwise, fraudsters could just rotate between tactics to evade detection. 

Continual learning performs well under these constraints. First, it updates existing models, so limited additional training is required. Second, updating the model can be done using only the most recent data, so there is no need to store all historical data indefinitely. Finally, continual learning is specifically designed to retain previous knowledge when learning from new data, so older modi operandi should still be detected. 

Despite all this, research on continual graph learning for AML is rare. Furthermore, current experiments and benchmarks in literature lack an in-depth discussion on the effect of the many choices underlying the applied continual learning framework. Therefore, this work sets out to answer the following research questions:
\begin{enumerate}[label=\textbf{RQ\arabic*}, ref=RQ\arabic*]
    \item \label{RQ lit} What is the current state of the literature on continual graph learning for anti-money laundering?
    \item \label{RQ hyp} What is the impact of the hyperparameters of the GNN and continual learning methods on performance and forgetting?
    \item \label{RQ DW} What is the impact of depth and width of the GNN on performance and forgetting?
    \item \label{RQ method} Which methods are best suited to overcome catastrophic forgetting for anti-money laundering? 
\end{enumerate}

To answer these questions, we conduct an in-depth literature review, summarising the current research on GNNs for AML, continual learning for financial fraud detection, and previous work on the effect of the involved hyperparameters. This review of the literature is complemented by an extensive experimental study to analyse the performance and forgetting of AML network methods on two open-source data sets. 
The contributions of our work are, hence, as follows:
\begin{itemize}
    \item We present an in-depth review of the current state-of-the-art in continual graph learning for fraud detection;
    \item We introduce and investigate the implications of continual graph learning on anti-money laundering, for edge as well as node classification;
    \item We present the result of extensive experiments on two open-source AML data sets and analyse the effects of various choices on performance. 
\end{itemize}
The code of the presented experiments is publicly available on github\footnote{\url{https://github.com/VerbekeLab}} to facilitate peer researchers and practitioners to replicate and extend the reported results. 

The remainder of this paper is organised as follows. Preliminary theory on graphs, graph neural networks and continual learning is discussed in Section~\ref{sec: preliminaries}. A review of the literature is presented in Section~\ref{sec:relatedwork}. Section~\ref{sec:methodology} presents the experimental methodology, with results and discussion provided in Section~\ref{sec:resultsanddiscussion}. Section~\ref{sec:conclusion} concludes this work and presents directions for future research.   

\section{Preliminaries}
\label{sec: preliminaries}
\subsection{Graphs}
A graph $G(V,E)$ is defined via two sets, $V$ and $E$. The elements of set $V=\{v_1, \ldots, v_n\}$ represent the nodes in the graph, while set $E \subset V\times V$ represents the edges that connect the nodes. An edge between node $i$ and $j$ is denoted as $e_{ij}$. In this work, we consider homogeneous networks. 
It is assumed that nodes are assigned a vector $x_i \in \R^m$ with feature values. The matrix containing all feature vectors is denoted by $X\in \R^{n\times m}$. 

\subsection{Graph Neural Networks} 
The initial idea of deep learning on graphs was introduced by \citet{scarselli2005graph} and \citet{scarselli2008graph}, based on message passing. The idea of message passing is still present in GNNs, where a node's representation is updated iteratively based on the node's neighbours. 

Formally, given a graph $G(V,E)$, graph neural networks (GNNs) construct the representation of node $i$ at layer $l$, denoted by $h_i^{(l)}$, as
\begin{equation}
h_i^{(l)} =  \phi^{(l-1)}\left( h_i^{(l-1)}, \sum_j \hat{A}_{ij} \psi^{(l-1)}\left(h_i^{(l-1)}, h_j^{(l-1)}\right) \right),  
\label{eq: def GNN}
\end{equation}
where $\phi^{(l)}$, and $\psi^{(l)}$ are layer-dependent functions, and $\hat{A}_{ij}$ is the normalized adjacency matrix, including self-loops. Most of the time, the initial embedding is set equal to the node features, $h_i^{(0)} = x_i$.

The most widely adopted graph neural networks are Graph Convolutional Networks~(GCN)~\citep{kipf2017semisupervised}, Graph SAmple and aggreGatE~(GraphSAGE)~\citep{NIPS2017_5dd9db5e}, Graph ATtention network~(GAT)~\citep{veličković2018graph} and  Graph Isomorphism Networks~(GIN)~\citep{xu2019powerful}. 
GCNs introduced by \citet{kipf2017semisupervised} aggregate neighbourhood information based on convolutions. \citet{NIPS2017_5dd9db5e} extends on this idea by introducing GraphSAGE resulting into an inductive method. \citet{veličković2018graph} introduces attention mechanisms using GAT, allowing the distinction between important and less important neighbours in the network. Finally, \citet{xu2019powerful} introduced GIN, relying on the Weisfeiler-Lehman graph isomorphism test to come to a more versatile version of GNNs. 

\subsection{Continual Learning}
In continual learning, a model needs to sequentially learn disjoint tasks $\mathcal{T}=\{\mathcal{T}_1, \ldots, \mathcal{T}_K\}$~\citep{EWC, https://doi.org/10.1002/widm.1526}. Specific observations are provided with each task $\mathcal{T}_i$, while access to data of previous tasks is often limited or even prohibited. Each task has its corresponding feature set $X_i$, and task-specific label $y_l\in \mathcal{Y}_i$, with label set $\mathcal{Y}_i= \{y^1, \ldots, y^{c_i}\}$, where $c_i$ represents the number of classes in task $\mathcal{T}_i$. Sometimes, a specific assumption is made that task data is provided as $(\mathcal{X},\mathcal{Y}, \mathcal{D}_\mathcal{C})$ with $\mathcal{D}_\mathcal{C}$ the underlying distribution, also called context set~\citep{de2021continual}.

Depending on the available information and the format of the tasks provided, four different continual learning settings are discerned, i.e., task-incremental, domain-incremental, class-incremental and time-incremental learning~\citep{vandeven2022vandeven,ko2024beginextensivebenchmarkscenarios}. The first three are well-established in continual learning~\citep{vandeven2022vandeven}. \textbf{Task-incremental learning} consists of a sequence of distinct tasks to be learned, where the model knows which task is currently presented, even at test time. \textbf{Domain-incremental learning} describes the scenario in which the problem is the same, but the distribution of the tasks shifts. Here, no information about the task is provided at test time. In \textbf{class-incremental learning}, a growing number of classes are provided with each new task, but no task information is provided. Hence, the methods should also be able to learn to distinguish the current task that is provided. \textbf{Time-incremental learning} encompasses problems where data is provided in streaming format, and where the distribution might shift over time. Some work considers this to be a separate setting~\citep{ko2024beginextensivebenchmarkscenarios}, while it can also be seen as a specific case of domain-IL. 

To mitigate catastrophic forgetting, different methods have been developed, broadly classified into three categories, i.e., replay, regularisation-based and parameter isolation~\citep{de2021continual}. \textbf{Replay} methods preserve some historical observation - either real or synthetic - to revisit when training on new tasks. \textbf{Regularization-based} methods use heuristics to determine the important weights in the neural network and to penalize changing these weights more when learning new tasks. Finally, \textbf{parameter isolation}, also referred to as architecture-based, reserves specific weights to be updated on specific tasks. This can be done by freezing part of the network, or extending the neural network for new tasks.  

Specific evaluation metrics are used to evaluate continual learning methods. The most widely used are average accuracy, average forgetting and forward transfer~\citep{ko2024beginextensivebenchmarkscenarios, https://doi.org/10.1002/widm.1526}. These evaluate the performance after learning all tasks. \textbf{Average accuracy} is the average of accuracy over the tasks, while average forgetting assesses the degradation of accuracy over the tasks. \textbf{Average forgetting} compares the accuracy of a task after training on that tasks to the accuracy after learning on all tasks. Negative forgetting is sometimes also called backward transfer. \textbf{Forward transfer}, also called zero-shot learning, is the increase in accuracy when using the model trained on previous tasks, compared to random initialization~\citep{GEM, https://doi.org/10.1002/widm.1526}.

\subsection{Continual Graph Learning}
\begin{figure}
    \centering
    \includegraphics[width=0.7\linewidth]{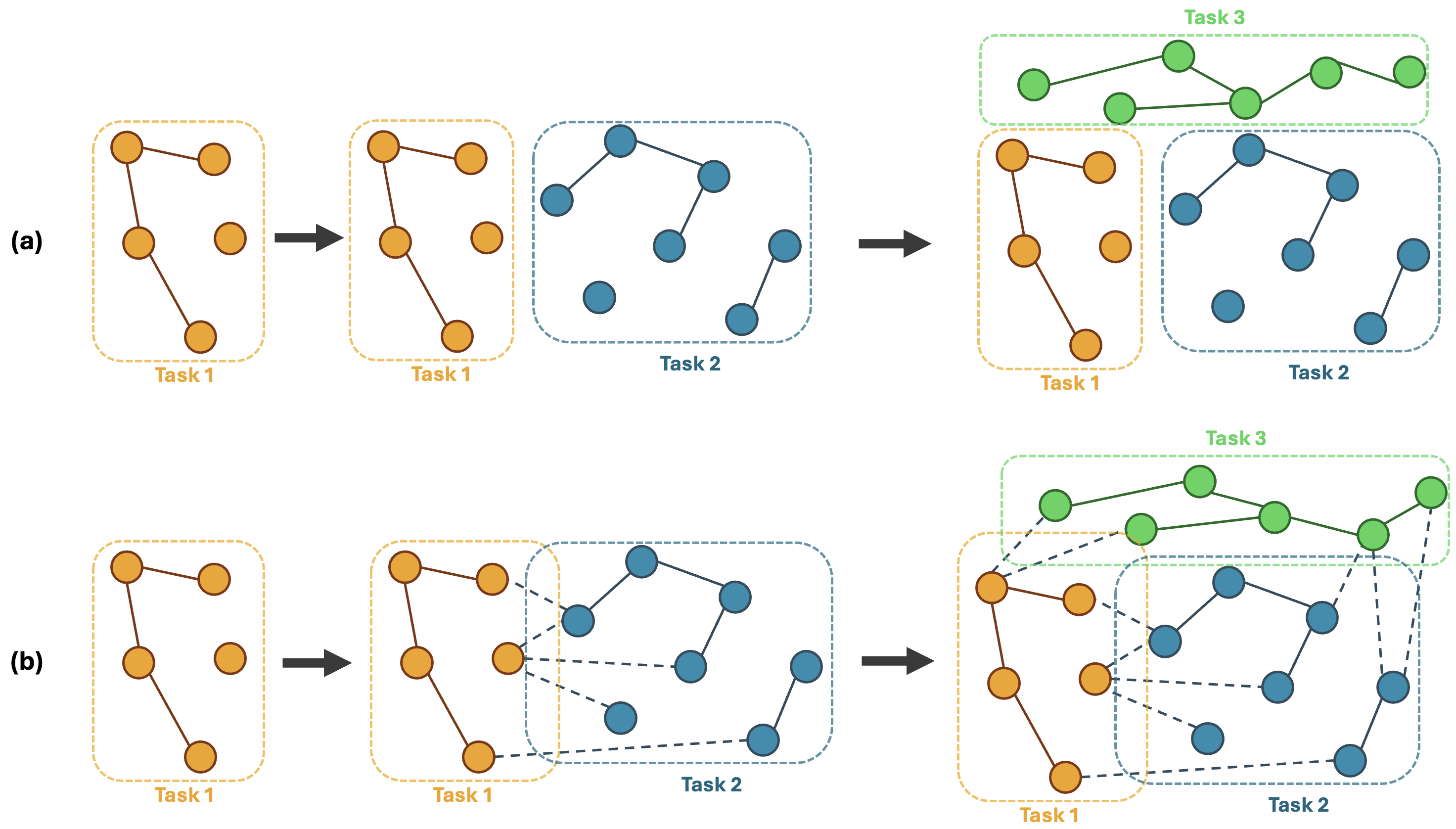}
    \caption{Two assumptions in continual graph learning for node-based learning. Either a separate network is provided for each task (a), or the network grows over time (b).}
    \label{Figure1}
\end{figure}

When using network data, additional considerations come into play for continual graph learning, since observations over different tasks may be connected in the network. As a result, information from previous tasks can still be leveraged when training the current tasks due to these inter-task connections. Suppose a classification problem, where each node is part of only one task~\citep{ko2024beginextensivebenchmarkscenarios}. The inter-task connections often fall within one of two categories, as visualised in Figure~\ref{Figure1}. 

A first category is where each task consists of an independent network with no links to nodes in previous tasks~\citep{li2022forgettingpreventioncrossregionalfraud, ko2024beginextensivebenchmarkscenarios}, so no inter-task connections are present. This can occur for two reasons. First, there are tasks which inherently do not have inter-task connections. This is often the case in graph classification~\citep{carta2021catastrophicforgettingdeepgraph}, but also occurs when every sample in a task is a separate network. Second, separate tasks can also occur by design, where additional restrictions are put on the network by removing inter-task connections. 

The second type of interaction is more prevalent, and involves a network that grows over time~\citep{CGNN,zhang2024continuallearninggraphschallenges,10026151,yuan2023continualgraphlearningsurvey, CGLB, ER-GNN}. With each new task, new nodes are added to the network. Some nodes in the new task have connections to nodes of previous tasks. Hence, special care is needed if we assume that not all data is available from previous tasks.

\begin{figure}
    \centering
    \includegraphics[width=0.7\linewidth]{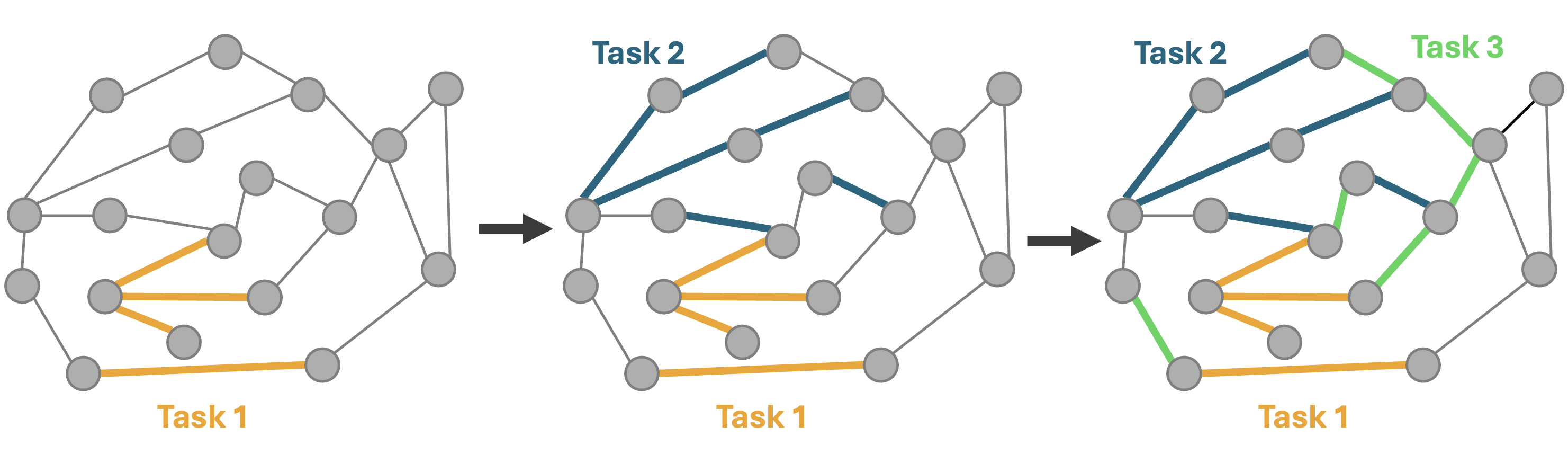}
    \caption{Visualisation of edge-based continual learning, where the network stays fixed, but the edge labels gradually become known.}
    \label{Figure2}
\end{figure}

In this work, we also evaluate a continual learning setting where the network is fixed, but where the labels change over time, as shown in Figure~\ref{Figure2}. Here, the nodes are shared across tasks. This corresponds closely to real-life AML settings, where clients are monitored continuously. A banking client can start laundering money only after holding an account at the bank for a couple of years. 

\section{Literature Review}
\label{sec:relatedwork}
In the past years, deep representation learning has gained increased adoption for AML, although remaining under-explored~\citep{deprez2024aml}. In the same vein, the field of continual learning is mature, but less attention has been given to continual graph learning~\citep{CGLB, ko2024beginextensivebenchmarkscenarios}. 
To answer \ref{RQ lit}, this section provides an overview of that literature. An overview of the relevant literature on continual learning is given in Table~\ref{tab:summary literature}

\subsection{Graph Neural Networks for Anti-Money Laundering}
\label{subsec:GNN AML}
\citet{alarab2020competence} present experiments with a GCN~\citep{kipf2017semisupervised} to construct meaningful network embeddings for AML. Cases are classified by combining the embedding with original transaction features.
\citet{inspectionL} introduced inspection-L, which applies a GIN~\citep{xu2019powerful} and uses Deep graph infomax to find predictive node embeddings by comparing the embeddings of the real network to that of a corrupted network. 
The application of GNNs is extended by \citet{jin2022heterogeneous}. First, a heterogeneous interaction network is constructed to extract additional features. Then, the authors apply GCN~\citep{kipf2017semisupervised}, GraphSAGE~\citep{NIPS2017_5dd9db5e} and GIN~\citep{xu2019powerful} on the homogeneous network to arrive at network embeddings augmented with the features from the heterogeneous network, so as to obtain the final predictions. 

As money laundering typically occurs over a longer period of time, research has also aimed at extending GNNs to a spatio-temporal setting. This is often done by constructing embeddings on snapshots that are fed into a deep learning method for time series analysis. 
The work by \citep{xia2021novel} feeds the embeddings coming from a GCN~\citep{kipf2017semisupervised} to an LSTM, while \citep{DynGraphTrans} applies a transformer model to the embedding vectors of the different snapshots. 

Furthermore, GNNs have been used for AML anomaly detection. 
\citet{cardoso2022laundrograph} use GAT~\citep{veličković2018graph} for link prediction between nodes in the network. These predictions are compared to the real links in the network, leading to anomaly scores that indicate suspicious transactions. 

\subsection{Continual (Graph) Learning Methods}
\label{subsec: CGL}
When training (graph) neural networks in the classic way, it is assumed that the data is identically distributed, often with the possibility to shuffle it before using it to train the model~\citep{PARISI201954}. However, data is typically not identically distributed in many real-life applications, where data becomes available in streams over time. An added difficulty arises when there is also a distributional shift in the tasks to learn. This can lead to catastrophic forgetting~\citep{catastrophicForgetting, MCCLOSKEY1989109, goodfellow2015empiricalinvestigationcatastrophicforgetting}, where updating the model with new information leads to interference with earlier-acquired knowledge. The trade-off between the ability to retain old knowledge while also processing new information is called the stability-plasticity dilemma~\citep{de2021continual,10599804,wang2024comprehensive}.

Continual learning, which was introduced to mitigate catastrophic forgetting, has mostly been applied for image classification~\citep{https://doi.org/10.1002/widm.1558}. One starts with a couple of images to classify, e.g., cat versus dog. The classes are then extended where the model also needs to be able to distinguish between, e.g., cars and planes. When learning this second task, the model should retain its ability to distinguish between cats and dogs. 

Given the origin of the field, many of the proposed methods are also evaluated on image classification~\citep{iCaRL, GEM, EWC, SI, lwf, packnet, de2021continual}. The most frequently used data sets are MNIST, CIFAR100 and ImageNet, and these also serve as the main data sets in benchmarking studies~\citep{de2021continual, carta2021catastrophicforgettingdeepgraph, vandeven2022vandeven, 10599804}. 

Continual learning methods are classified in three categories, i.e., replay, regularisation-based and architecture-based~\citep{de2021continual, https://doi.org/10.1002/widm.1558}, whereas hybrid methods also exist. 

\textbf{Replay methods} rely on a memory buffer for retaining a subset of data from previous tasks, called exemplars. These methods assume that resources are available to store previous data. 
Replay emerged in reinforcement learning~\citep{CLEAR}, where past transitions are replayed for better learning on a single task, while replay in continual learning is used to retain information across tasks.
Care should be taken when adopting such methods with regards to regulation. Due to privacy concerns, not all data is allowed to be stored indefinitely. Additionally, it is possible that these methods overfit on the few exemplars that are kept in memory~\citep{CGNN}. 

Pure replay methods only specify a buffer size $\mathcal{B}$, and randomly assign $\mathcal{B}/k$ observations from the current task to be replayed during new tasks. This random selection is also applied by GEM~\citep{GEM} to select their exemplar set. iCarl~\citep{iCaRL}, on the other hand, uses smart allocation by selecting exemplars that best preserve the average feature vector of that task. 

Additionally, replay data can be generated synthetically. One way of doing this is proposed by \citet{lwf} for their method Learning without Forgetting~(LwF) in a task-IL setting. The authors assume that each task has its own output head. As data on previous tasks is not available, LwF takes the data of the current task, and makes predictions for the output heads of the previous tasks as well. This gives a new `ground truth' for previous tasks, to which output is compared to during training, to keep the output on previous tasks stable. 

Replay methods have also been extended to graph learning. 
\citet{ER-GNN} use the \textit{mean of features} of selecting exemplars, and extend it for graph data by selecting exemplars based on the mean embedding, based on coverage maximisation and based on influence maximisation. \citet{CGNN} employ a two-step sampling approach where first the network is divided into clusters, after which nodes are selected within each cluster based on an importance metric. 

For the class-IL setting, \citet{CAT} proposes CaT, which first selects a subset of nodes randomly, and then uses a structure-free graph condensation method \citep{zheng2023structure} to align the mean of latent features in the subset of nodes by training the input node features as weights.

\textbf{Regularisation-based methods} limit the updates to weights in the (graph) neural network. Determining which weights can change and by how much leads to a trade-off between the stability and plasticity of the model~\citep{PARISI201954}.

Regularisation-based methods often capture which weights are important for previous tasks, and limit how much these can be changed. This is done by introducing additional terms in the loss function. EWC~\citep{EWC} uses the elements in the Fisher information matrix to express which weights were important for previous tasks. As mentioned by \citet{SI}, EWC only makes point estimates of the importance using the diagonal elements of the Fisher information matrix. They put forward SI~\citep{SI}, where the importance of the weights are continuously calculated throughout the training process.

MAS~\citep{mas} on the other hand, is based on the gradient of the squared $l_2$-norm of the learning function output, making it applicable to unlabelled data as well. Simplifications of the importance calculations are given in case all layers use a ReLU activation function. 

Regularisation can also be approached by altering the gradient before updating the weights. Using the exemplars in GEM, \citet{GEM} project the gradient on the span of the gradients of the previous tasks. The authors claim that this does not increase the loss on older tasks when updating the weights for the new task. 

\citet{twp} extends regularisation to network data with TWP. Two importance scores are included, i.e., a task-related one similar to EWC and a topology-related one based on the attention mechanism in graph attention networks (GAT)~\citep{veličković2018graph}. 
Similarly, to mitigate the problem of overfitting on the exemplars, \citet{CGNN} applies the same idea of EWC to GNNs. The authors use the diagonal elements of the Fisher information matrix to regularize the updates of important weights in the GNN. 

\textbf{Architecture-based methods}, as their name suggests, alter the architecture of the (graph) neural network based on the tasks. Specific parameters can be isolated to be fine-tuned, or the architecture itself can be extended for new classes, e.g., have separate output heads for each task~\citep{lwf}. Here, knowledge of the current task must be provided. Some methods also assume that the total number of tasks is known upfront. This limits their adoption in practice. 

\citet{vandeven2022vandeven} mention the use of entirely separate output layers or networks to learn each task. In the task-IL setting, \citet{lwf} initializes a new output layer for each new task. Other methods include the usage of gating. XdG~\citep{XdG} introduced a context-dependent gating signal, to have sparsely connecting, mostly non-overlapping parts of the network trained on the different tasks. 

Similar to gating, PackNet~\citep{packnet} and piggyback~\citep{piggyback} apply task-specific masks to set some weights in the neural network equal to 0. When learning a task, \citet{packnet} train the network and then prune it. The remaining weights are then fine-tuned in a second, shorter training round. These weights are fixed and the pruned weights are made available for the next task. The main drawback is that less capacity is available for training on the next task, meaning that PackNet can only learn a limited number of tasks. This is mitigated in the work by \citet{piggyback}. Here, the network weights are fixed and the task-specific masks are learned. The authors use gradient-based learning to obtain real-valued masks, which are then converted to binary masks using a fixed threshold. The main drawback here, next to the need to know which task is considered, is that piggyback requires a pre-trained network. The performance of piggyback is highly dependent on this pre-training step~\citep{piggyback}, and this might even not be available depending on the application. 

\begin{table}[]
\tiny
\centering
\Rotatebox{90}{%
    \begin{tabular}{|l|cc|cccc|cccccccccccccc|}
    \toprule
        Paper & Graph Data & Fraud & GCN & GraphSAGE & GAT & GIN &
         Replay&iCarl & EWC & MAS & GEM & TWP & LwF & ER-GNN & PackNet & Piggyback & HAT & CaT & PI-GNN & CGNN\\\midrule
        \citet{CGLB} & \ding{51} & \textcolor{gray!25}{\ding{55}} & \ding{51} & \textcolor{gray!25}{\ding{55}} & \textcolor{gray!25}{\ding{55}} & \textcolor{gray!25}{\ding{55}} &  \textcolor{gray!25}{\ding{55}}&\textcolor{gray!25}{\ding{55}} & \ding{51} &\ding{51}&\ding{51}&\ding{51}&\ding{51}&\ding{51}&\ding{51}&\ding{51}&\ding{51}&\ding{51}&\ding{51}&\ding{51} \\
         \citet{ko2024beginextensivebenchmarkscenarios} & \ding{51} & \textcolor{gray!25}{\ding{55}} & \ding{51} & \textcolor{gray!25}{\ding{55}} & \textcolor{gray!25}{\ding{55}} & \textcolor{gray!25}{\ding{55}} &  \textcolor{gray!25}{\ding{55}}&\textcolor{gray!25}{\ding{55}} &\ding{51}&\ding{51}&\ding{51}&\ding{51}&\ding{51}&\ding{51} & \textcolor{gray!25}{\ding{55}} & \textcolor{gray!25}{\ding{55}} & \textcolor{gray!25}{\ding{55}} & \textcolor{gray!25}{\ding{55}} & \textcolor{gray!25}{\ding{55}} & \textcolor{gray!25}{\ding{55}} \\
         \citet{10599804} & \textcolor{gray!25}{\ding{55}} & \textcolor{gray!25}{\ding{55}} & \textcolor{gray!25}{\ding{55}} & \textcolor{gray!25}{\ding{55}} & \textcolor{gray!25}{\ding{55}} & \textcolor{gray!25}{\ding{55}} &  \ding{51}&\ding{51} & \ding{51} & \textcolor{gray!25}{\ding{55}} & \ding{51} & \textcolor{gray!25}{\ding{55}} & \ding{51} & \textcolor{gray!25}{\ding{55}} & \textcolor{gray!25}{\ding{55}} & \textcolor{gray!25}{\ding{55}} & \textcolor{gray!25}{\ding{55}} & \textcolor{gray!25}{\ding{55}} & \textcolor{gray!25}{\ding{55}} & \textcolor{gray!25}{\ding{55}} \\
         \citet{de2021continual} & \textcolor{gray!25}{\ding{55}} & \textcolor{gray!25}{\ding{55}} & \textcolor{gray!25}{\ding{55}} & \textcolor{gray!25}{\ding{55}}& \textcolor{gray!25}{\ding{55}}&\textcolor{gray!25}{\ding{55}} &  \textcolor{gray!25}{\ding{55}}&\ding{51} & \ding{51} & \ding{51} &\ding{51} & \textcolor{gray!25}{\ding{55}} &\ding{51} & \textcolor{gray!25}{\ding{55}}&\ding{51}&\textcolor{gray!25}{\ding{55}}&\ding{51}& \textcolor{gray!25}{\ding{55}} &\textcolor{gray!25}{\ding{55}}&\textcolor{gray!25}{\ding{55}}\\
         \citet{10026151} & \ding{51} & \textcolor{gray!25}{\ding{55}} & \ding{51}&\ding{51}&\ding{51}&\ding{51}&  \textcolor{gray!25}{\ding{55}}&\textcolor{gray!25}{\ding{55}}&\textcolor{gray!25}{\ding{55}}&\textcolor{gray!25}{\ding{55}}&\textcolor{gray!25}{\ding{55}}&\ding{51}&\textcolor{gray!25}{\ding{55}}&\ding{51}&\textcolor{gray!25}{\ding{55}}&\textcolor{gray!25}{\ding{55}}&\textcolor{gray!25}{\ding{55}}&\textcolor{gray!25}{\ding{55}}&\textcolor{gray!25}{\ding{55}}&\ding{51} \\
         \citet{zhang2024continuallearninggraphschallenges} & \ding{51} & \textcolor{gray!25}{\ding{55}} & \textcolor{gray!25}{\ding{55}}&\textcolor{gray!25}{\ding{55}}&\textcolor{gray!25}{\ding{55}}&\textcolor{gray!25}{\ding{55}}&  \textcolor{gray!25}{\ding{55}}&\textcolor{gray!25}{\ding{55}}&\ding{51}&\ding{51} & \textcolor{gray!25}{\ding{55}} & \ding{51} & \textcolor{gray!25}{\ding{55}} & \ding{51} & \textcolor{gray!25}{\ding{55}}&\textcolor{gray!25}{\ding{55}}&\textcolor{gray!25}{\ding{55}}&\ding{51}&\ding{51} & \textcolor{gray!25}{\ding{55}}\\
         \citet{Wei_2024_CVPR} & \ding{51} & \textcolor{gray!25}{\ding{55}} & \ding{51} & \textcolor{gray!25}{\ding{55}}&\textcolor{gray!25}{\ding{55}}&\textcolor{gray!25}{\ding{55}}& \textcolor{gray!25}{\ding{55}}&\textcolor{gray!25}{\ding{55}}&\textcolor{gray!25}{\ding{55}}&\textcolor{gray!25}{\ding{55}}&\textcolor{gray!25}{\ding{55}}&\textcolor{gray!25}{\ding{55}}&\textcolor{gray!25}{\ding{55}}&\textcolor{gray!25}{\ding{55}}&\textcolor{gray!25}{\ding{55}}&\textcolor{gray!25}{\ding{55}}&\textcolor{gray!25}{\ding{55}}&\ding{51}&\textcolor{gray!25}{\ding{55}}&\textcolor{gray!25}{\ding{55}}\\
         \citet{CGNN} & \ding{51} & \ding{51} & \textcolor{gray!25}{\ding{55}} & \ding{51} & \textcolor{gray!25}{\ding{55}}&\textcolor{gray!25}{\ding{55}}& \textcolor{gray!25}{\ding{55}}&\textcolor{gray!25}{\ding{55}}&\textcolor{gray!25}{\ding{55}}&\textcolor{gray!25}{\ding{55}}&\textcolor{gray!25}{\ding{55}}&\textcolor{gray!25}{\ding{55}}&\textcolor{gray!25}{\ding{55}}&\textcolor{gray!25}{\ding{55}}&\textcolor{gray!25}{\ding{55}}&\textcolor{gray!25}{\ding{55}}&\textcolor{gray!25}{\ding{55}}&\textcolor{gray!25}{\ding{55}}&\textcolor{gray!25}{\ding{55}}&\ding{51}\\
         \citet{hemati2022continuallearningunsupervisedanomaly} & \textcolor{gray!25}{\ding{55}} & \ding{51} & \textcolor{gray!25}{\ding{55}}&\textcolor{gray!25}{\ding{55}}&\textcolor{gray!25}{\ding{55}}&\textcolor{gray!25}{\ding{55}} &  \ding{51}&\textcolor{gray!25}{\ding{55}}&\ding{51} & \textcolor{gray!25}{\ding{55}}&\textcolor{gray!25}{\ding{55}}&\textcolor{gray!25}{\ding{55}}&\textcolor{gray!25}{\ding{55}}&\textcolor{gray!25}{\ding{55}}&\textcolor{gray!25}{\ding{55}}&\textcolor{gray!25}{\ding{55}}&\textcolor{gray!25}{\ding{55}}&\textcolor{gray!25}{\ding{55}}&\textcolor{gray!25}{\ding{55}}&\textcolor{gray!25}{\ding{55}} \\
         \citet{Zhang_Cheng_Yang_Ouyang_Wu_Zheng_Jiang_2024} & \ding{51} &\ding{51} & \textcolor{gray!25}{\ding{55}}&\textcolor{gray!25}{\ding{55}}&\ding{51}&\textcolor{gray!25}{\ding{55}} & \textcolor{gray!25}{\ding{55}}&\textcolor{gray!25}{\ding{55}}&\textcolor{gray!25}{\ding{55}}&\ding{51} &\textcolor{gray!25}{\ding{55}}&\textcolor{gray!25}{\ding{55}}&\textcolor{gray!25}{\ding{55}}&\ding{51} &\textcolor{gray!25}{\ding{55}}&\textcolor{gray!25}{\ding{55}}&\textcolor{gray!25}{\ding{55}}&\textcolor{gray!25}{\ding{55}}&\textcolor{gray!25}{\ding{55}}&\ding{51}\\
         \citet{li2022forgettingpreventioncrossregionalfraud} & \ding{51} & \ding{51} & \textcolor{gray!25}{\ding{55}}&\textcolor{gray!25}{\ding{55}}& \ding{51} & \textcolor{gray!25}{\ding{55}} &  \textcolor{gray!25}{\ding{55}}&\textcolor{gray!25}{\ding{55}} &\ding{51} & \ding{51} & \ding{51} & \textcolor{gray!25}{\ding{55}} &\textcolor{gray!25}{\ding{55}} & \textcolor{gray!25}{\ding{55}} & \textcolor{gray!25}{\ding{55}}&\textcolor{gray!25}{\ding{55}}&\ding{51}&\textcolor{gray!25}{\ding{55}}&\textcolor{gray!25}{\ding{55}}&\textcolor{gray!25}{\ding{55}}\\
         \citet{vandeven2022vandeven} & \textcolor{gray!25}{\ding{55}}&\textcolor{gray!25}{\ding{55}}&\textcolor{gray!25}{\ding{55}}&\textcolor{gray!25}{\ding{55}}&\textcolor{gray!25}{\ding{55}}&\textcolor{gray!25}{\ding{55}}& \textcolor{gray!25}{\ding{55}}&\ding{51}&\ding{51}& \textcolor{gray!25}{\ding{55}} & \ding{51} & \textcolor{gray!25}{\ding{55}} & \ding{51} & \textcolor{gray!25}{\ding{55}}&\textcolor{gray!25}{\ding{55}}&\textcolor{gray!25}{\ding{55}}&\textcolor{gray!25}{\ding{55}}&\textcolor{gray!25}{\ding{55}}&\textcolor{gray!25}{\ding{55}}&\textcolor{gray!25}{\ding{55}}\\
         \citet{carta2021catastrophicforgettingdeepgraph} & \ding{51}&\textcolor{gray!25}{\ding{55}}&\textcolor{gray!25}{\ding{55}}&\textcolor{gray!25}{\ding{55}}&\textcolor{gray!25}{\ding{55}}&\textcolor{gray!25}{\ding{55}}&  \ding{51}&\textcolor{gray!25}{\ding{55}}&\ding{51}& \textcolor{gray!25}{\ding{55}}&\textcolor{gray!25}{\ding{55}}&\textcolor{gray!25}{\ding{55}}&\ding{51}& \textcolor{gray!25}{\ding{55}}&\textcolor{gray!25}{\ding{55}}&\textcolor{gray!25}{\ding{55}}&\textcolor{gray!25}{\ding{55}}&\textcolor{gray!25}{\ding{55}}&\textcolor{gray!25}{\ding{55}}&\textcolor{gray!25}{\ding{55}}\\
         \citet{ER-GNN} & \ding{51} & \textcolor{gray!25}{\ding{55}} &\textcolor{gray!25}{\ding{55}}&\textcolor{gray!25}{\ding{55}}&\ding{51}&\textcolor{gray!25}{\ding{55}}&\textcolor{gray!25}{\ding{55}}&\textcolor{gray!25}{\ding{55}}&\textcolor{gray!25}{\ding{55}}&\textcolor{gray!25}{\ding{55}}&\textcolor{gray!25}{\ding{55}}&\textcolor{gray!25}{\ding{55}}&\textcolor{gray!25}{\ding{55}}&\ding{51}&\textcolor{gray!25}{\ding{55}}&\textcolor{gray!25}{\ding{55}}&\textcolor{gray!25}{\ding{55}}&\textcolor{gray!25}{\ding{55}}&\textcolor{gray!25}{\ding{55}}&\textcolor{gray!25}{\ding{55}} \\
         \citet{LEBICHOT2024123445} & \textcolor{gray!25}{\ding{55}}&\ding{51}&\textcolor{gray!25}{\ding{55}}&\textcolor{gray!25}{\ding{55}}&\textcolor{gray!25}{\ding{55}}&\textcolor{gray!25}{\ding{55}}&\ding{51}&\textcolor{gray!25}{\ding{55}}&\ding{51}&\textcolor{gray!25}{\ding{55}}&\textcolor{gray!25}{\ding{55}}&\textcolor{gray!25}{\ding{55}}&\textcolor{gray!25}{\ding{55}}&\textcolor{gray!25}{\ding{55}}&\textcolor{gray!25}{\ding{55}}&\textcolor{gray!25}{\ding{55}}&\textcolor{gray!25}{\ding{55}}&\textcolor{gray!25}{\ding{55}}&\textcolor{gray!25}{\ding{55}}&\textcolor{gray!25}{\ding{55}}\\
         \citet{10.1145/3477314.3507113} & \ding{51} & \ding{51} & \textcolor{gray!25}{\ding{55}} & \ding{51} & \textcolor{gray!25}{\ding{55}} &\textcolor{gray!25}{\ding{55}} &\ding{51} & \textcolor{gray!25}{\ding{55}}&\textcolor{gray!25}{\ding{55}}&\textcolor{gray!25}{\ding{55}}&\textcolor{gray!25}{\ding{55}}&\textcolor{gray!25}{\ding{55}}&\textcolor{gray!25}{\ding{55}}&\textcolor{gray!25}{\ding{55}}&\textcolor{gray!25}{\ding{55}}&\textcolor{gray!25}{\ding{55}}&\textcolor{gray!25}{\ding{55}}&\textcolor{gray!25}{\ding{55}}&\textcolor{gray!25}{\ding{55}}&\textcolor{gray!25}{\ding{55}} \\
         \citet{Zhang_Cheng_Yang_Ouyang_Wu_Zheng_Jiang_2024} & \ding{51} & \ding{51} & \textcolor{gray!25}{\ding{55}} & \ding{51} & \ding{51} & \textcolor{gray!25}{\ding{55}} & \textcolor{gray!25}{\ding{55}} & \textcolor{gray!25}{\ding{55}} & \textcolor{gray!25}{\ding{55}} & \ding{51} & \textcolor{gray!25}{\ding{55}} & \textcolor{gray!25}{\ding{55}} & \textcolor{gray!25}{\ding{55}} & \ding{51} & \textcolor{gray!25}{\ding{55}} & \textcolor{gray!25}{\ding{55}}& \textcolor{gray!25}{\ding{55}}& \textcolor{gray!25}{\ding{55}}&\textcolor{gray!25}{\ding{55}}&\ding{51}\\
         \bottomrule
    \end{tabular}}
    \caption{Table summarising the main literature in continual (graph) learning.}
    \label{tab:summary literature}
\end{table}

\subsection{Continual Learning in Financial Fraud}
\label{subsec: CL fraud}
When implementing fraud detection in practice, the methods need to continuously monitor millions of transactions, which results in three main challenges. The first is that, given that fraud is evolving constantly~\citep{baesens2015fraud, VANVLASSELAER201538}, AML models should be updated to capture novel modi operandi. The second is constraints on computational resources. The large volume of transactions make retraining the model from scratch not always feasible given limited time and resources. The third challenge comes from the desire to retain knowledge on previously applied fraud tactics, because launderers could otherwise revert back to an older modus operandi to avoid detection. However, when fine-tuning the models, this information is lost if these tactics were not applied in the latest available data, resulting in catastrophic forgetting~\citep{catastrophicForgetting, MCCLOSKEY1989109, goodfellow2015empiricalinvestigationcatastrophicforgetting}.

Although continual learning can be used to mitigate these challenges, research on the adoption of continual learning for fraud detection is scarce~\citep{https://doi.org/10.1002/widm.1558}. Research on continual learning is mostly concerned with image recognition~\citep{https://doi.org/10.1002/widm.1558}. Few studies present fraud detection as the core problem~\citep{hemati2022continuallearningunsupervisedanomaly, Zhang_Cheng_Yang_Ouyang_Wu_Zheng_Jiang_2024, LEBICHOT2024123445, li2022forgettingpreventioncrossregionalfraud}. It often only appears as one of many data sets to which continual learning methods are applied~\citep{CGNN, 10.1145/3477314.3507113, ko2024beginextensivebenchmarkscenarios}. 

\citet{LEBICHOT2024123445} were among the first to quantify catastrophic forgetting for credit card fraud detection. They tested different replay strategies and EWC against fine-tuning the model. In their case, fine-tuning the model on new data seemed to be best at avoiding forgetting. 

\citet{hemati2022continuallearningunsupervisedanomaly} applied three strategies, sequential fine-tuning, EWC, and experience replay, for auditing financial payment records. They trained auto-encoders for anomaly detection and demonstrated that continual learning has the ability to detect distributional shifts. 

\citet{Zhang_Cheng_Yang_Ouyang_Wu_Zheng_Jiang_2024} introduced and applied a new method, called POCL, to medical insurance fraud detection. The authors rely on Temporal MAS to update the weights of their GNN. 

\citet{li2022forgettingpreventioncrossregionalfraud} extended continual graph learning to a case study on heterogeneous networks by introducing HTG-CFD. They apply replay and regularisation-based methods, where prototypes are constructed using the average attribute, and regularisation is done using Fisher information, inspired by EWC. HTG-CFD is constructed to transfer the fraud detection model across different regions to detect fraudulent transactions in a trade network. 

ContinualGNN~\citep{CGNN} is developed for streaming graphs to uncover new patterns over time. The authors have tested their method on the Elliptic data set. Although ContinualGNN did not perform best, the method has competitive performance. The main strength of this method is the strong reduction in training time compared to fully retraining the GNN with new data. 

\subsection{Benchmarks and Evaluation in Continual (Graph) Learning}
\label{subsec: benchmark}
\citet{de2021continual} are among the first to do an extensive benchmark study. While focusing only on task-incremental learning in classic continual learning, they implement a comprehensive benchmark both in terms of methods as well as data sets. These data sets all involve image classification. They conclude that architecture-based methods, particularly PackNet~\citep{packnet}, perform best, closely followed by memory replay. However, compared to memory replay, architecture-based methods do not suffer from privacy issues, since they do not require storing raw data.  

One of the first benchmark studies for continual graph learning was presented by \citet{carta2021catastrophicforgettingdeepgraph}. This study only considers graph classification, so no tests are done at node level. It is presented as an introductory benchmark experiment and it is quite limited in its scope. It involves three data sets and three continual learning strategies. These are naive replay, EWC~\citep{EWC} and LwF~\citep{lwf}. Hence, although the paper is meant to be a benchmark on networks, no strategies that were specifically developed for networks were tested. 

Two other notable benchmark studies for continual graph learning are \textit{Continual Graph Learning Benchmark~(CGLB)} by \citet{CGLB} and \textit{\textbf{Be}nchmarking \textbf{G}raph Cont\textbf{in}ual Learning~(BeGin)} by \citet{ko2024beginextensivebenchmarkscenarios}, both of which provide the full code suite to facilitate reproduction. The initial methods compared by \citet{CGLB} are EWC~\citep{EWC}, MAS~\citep{mas}, GEM~\citep{GEM}, TWP~\citep{twp}, LwF~\cite
{lwf} and ER-GNN~\citep{ER-GNN}. CGLB split continual graph learning in task-IL and class-IL and provide experiments for node-level and graph-level predictions. 

A more extensive benchmark is implemented by BeGin~\citep{ko2024beginextensivebenchmarkscenarios}. They make a fine-grained distinction between incremental settings, by considering task-IL, class-IL, domain-IL and time-IL. These settings are also introduced for link-level predictions, on top of the earlier introduced node-level and graph-level predictions. The continual learning methods are also extended. On top of the methods compared under CGLB, BeGin also includes PackNet~\citep{packnet}, Piggyback~\citep{piggyback}, HAT~\citep{HAT}, CaT~\citep{CAT}, PI-GNN~\citep{PI-GNN} and CGNN~\citep{CGNN}.

Both benchmarks analyse the performance of a range of methods, but pay less attention to the impact of the hyperparameters. One of the main shortcomings of both CGLB and BeGin is that all experiments are done only with GCN~\citep{kipf2017semisupervised} as backbone GNN. 

\subsection{Sensitivity to Hyperparameters}
\label{subsec: sensitivity}
Underlying every model is a suite of hyperparameter choices that impact model performance. These are often only briefly mentioned under hyperparameter tuning, or in the best case, papers apply limited parameter sensitivity tests. In this work, we make the effect of hyperparameters explicit. 

A key choice is the backbone GNN model. In the literature, GCN~\citep{kipf2017semisupervised} and GAT~\citep{veličković2018graph} are popular backbones, as indicated by \citet{yuan2023continualgraphlearningsurvey} and corroborated by our observation in Table~\ref{tab:summary literature}. Most continual graph learning methods are constructed to be backbone agnostic. One notable exception is TWP~\citep{twp}, which is specifically developed with GAT in mind. However, the authors provide proxies in case no attention mechanism is present. 

The architecture of the backbone GNN itself - both in terms of depth and width of the hidden layers - is also important since it influences the learning capacity and the length of transaction chains that can be captured. However, the specific impact of these choices have not been given much attention in literature. Studies on general continual learning by \citet{de2021continual} and \citet{mirzadeh2022architecturematterscontinuallearning} demonstrated that wide and shallow models generally perform better. However, when moving to continual graph learning, \citet{Wei_2024} demonstrated that for skeleton-based action recognition this does not always hold. 

Another import choice, next to the backbone, is the task definition. When considering human learning, it is hypothesized that the order of tasks is important for continual learning. \textit{Curriculum Learning}, coined by~\citet{10.1145/1553374.1553380}, determines that knowledge can be optimally acquired if tasks are learned in ascending order of difficulty. 

Previous work has performed task-order sensitivity analysis in continual learning. \citet{de2021continual} corroborated earlier work of~\citet{nguyen2019understandingcatastrophicforgettingcontinual} by showing that, in a general continual learning setting, methods exhibit order-agnostic behaviour. The authors test different setups, including an \textit{easy to hard}, a \textit{hard to easy}, and a random ordering of tasks. 

The work by \citet{packnet}, on the other hand, showed that for their method PackNet, the order does matter. They found that learning tasks from hardest to easiest actually gave better results. We cannot generalise this finding, however, since this is method-specific. The capacity of PackNet to incorporate novel information drops as the available free parameters decrease with each task. 

In the field of continual graph learning, \citet{zhao2024agale} introduced a randomly generated class appearance order to simulate the random class emergence in real world for multi-label continual graph learning. \citet{Wei_2024} focused on evaluating the task-order and class-order sensitivity in the context of continual graph learning for skeleton-based action recognition. The authors show that task-order robustness does not necessarily imply class-order robustness. 

\section{Methodology}
\label{sec:methodology}
\begin{figure}
    \centering
    \includegraphics[width=0.9\linewidth]{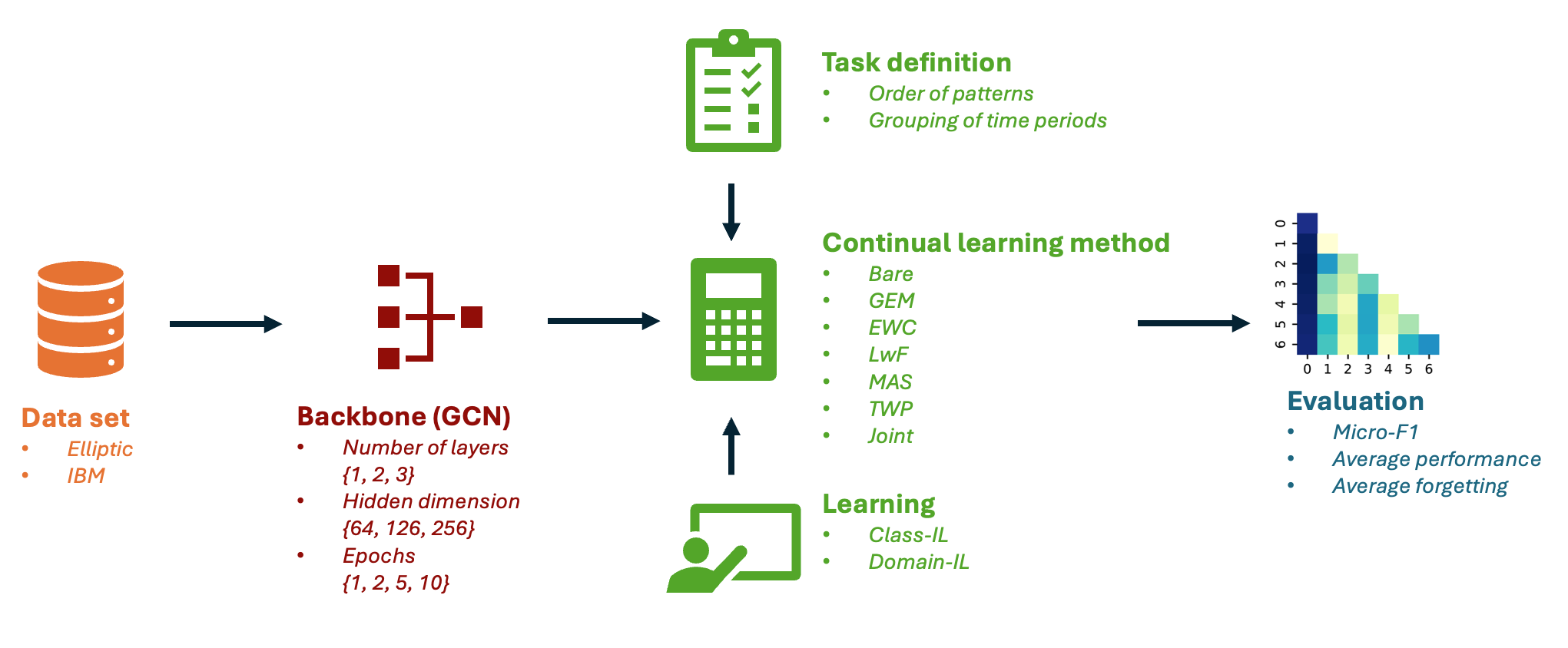}
    \caption{Visualisation of the experiment's pipeline and the hyperparameter choices for evaluation.}
    \label{Figure3}
\end{figure}

To answer the research questions, we set up a pipeline in which we vary the different hyperparameters~(\ref{RQ hyp}), the architecture of the GNN~(\ref{RQ DW}) and the continual learning methods~(\ref{RQ method}) to see their effect on performance and forgetting. 
Figure~\ref{Figure3} illustrates the full pipeline of our experiments, including the value of the hyperparameters.

The experiments presented in this paper are an extension of the BeGin framework that was introduced by~\citet{ko2024beginextensivebenchmarkscenarios}. The extended repository is made available on github\footnote{\url{https://github.com/VerbekeLab}}.

\subsection{Data}
Two data sets are used in the experiments that are widely used in AML~\citep{deprez2024aml}, i.e., the IBM AML data set~\citep{altman2024realistic} and the Elliptic data set~\citep{elliptic,weber2019anti}. As will be discussed below, this work presents AML both as an edge classification (IBM) as well as a node classification (elliptic) problem. 

\begin{figure}
    \centering
    \includegraphics[width=0.9\linewidth]{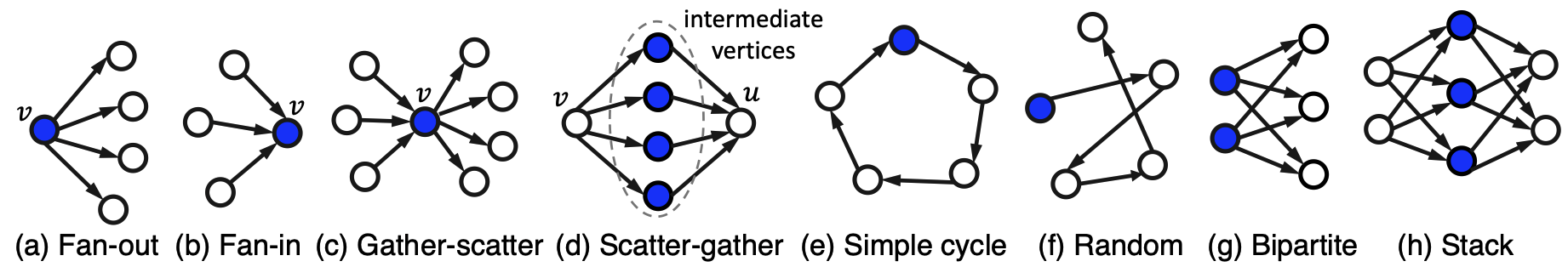}
    \caption{The different money laundering patterns as defined by \citet{altman2024realistic}.}
    \label{Figure4}
\end{figure}

The IBM AML data set~\citep{altman2024realistic} contains synthetic transaction data. The simulations are done via a virtual multi-agent virtual world. These agents can be banks, individuals or companies, with payments by individuals and companies. In this virtual world, some agents are said to be malicious. For those agents, the simulations include money laundering transactions. \citet{altman2024realistic} model eight different money laundering patterns, i.e., fan-in, fan-out, bipartite, stack, random, cycle, scatter-gather and gather-scatter, as illustrated in Figure~\ref{Figure4}. 

We select the HI-Small data set to use for our evaluations, taking the agents as nodes and the transactions as edges. This results in a network with 515~080 nodes and 5~078~345 edges. For this data set, we perform edge classification. As with most fraud data sets, the class distribution is highly imbalanced, with only $0.1\%$ of transactions involving money laundering. Most of the money laundering transactions are, however, not classified under a specific pattern. Table~\ref{tab:IBM distribution} gives a detailed view on the distribution of the class labels. We will discard the \textit{not classified} labels in our experiments, since we have no control over the specific patterns they constitute. This is also a practice observed in other research when the specific patterns are classified~\citep{altman2024realistic,egressy2024provably}.

\begin{table}[]
    \centering
    \begin{tabular}{|l|r|}
    \toprule
        Type of pattern & Transactions    \\
    \midrule
        Fan-out         & 342           \\
        Fan-in          & 318            \\
        Gather-scatter  & 716           \\
        Scatter-Gather  & 626         \\
        Cycle           & 287          \\
        Random          & 191         \\
        Bipartite       & 263         \\
        Stack           & 466        \\
        Not classified  & 1~968      \\
        Total money laundering & 5~177 \\
        \midrule
        Total number of transactions & 5~078~345 \\
    \bottomrule
    \end{tabular}
    \caption{The distribution of the different types of money laundering transaction patterns for the HI-Small data sets.}
    \label{tab:IBM distribution}
\end{table}

The Elliptic data set~\citep{elliptic,weber2019anti} contains real-world Bitcoin transactions, grouped in 49 different time intervals. The network consists of 20~3769 nodes and 234~355 edges, where nodes represent transactions and edges indicate that the receiver of the first transaction was the sender of the second. For this data set, we perform node classification. The data set includes 166 pre-calculated numerical features — 94 transaction-specific features and 72 aggregated features summarizing a node’s neighbours. The data contains only 4~545 illicit transactions ($2\%$), again making the label distribution highly imbalanced.  Although these labels do not specifically concern money laundering, we use this data set as it has found wide adoption in the AML literature~\citep{weber2019anti, deprez2024aml, alarab2020competence,xia2021novel, mohan2022improving, sun2022game, li2022blockchain, li2022transactional}, and therefore facilitates comparison with prior experimental results. 

Additionally, the Elliptic data set provides a well-suited case-study for continual learning~\citep{10.1145/3477314.3507113, CGNN}. As mentioned by \citet{weber2019anti}, there was a sudden closure of a dark market at time step 43. This caused all methods to perform poorly, due to the sudden shift in feature distribution of the illicit cases. This abrupt change in a real-world data set is ideal for getting a deeper understanding on continual learning methods.  

\subsection{Backbone Graph Neural Network}
In line with previous benchmarks~\citep{CGLB, ko2024beginextensivebenchmarkscenarios}, we use the GCN~\citep{kipf2017semisupervised} as backbone. The GCN layer-wise propagation is defined for the whole network at once as:
\begin{equation}
    H^{(l+1)} = \sigma\left( \tilde{D}^{-\frac{1}{2}}\tilde{A}\tilde{D}^{-\frac{1}{2}} H^{(l)} W^{(l)} \right),
    \label{eq:GCN}
\end{equation}
with $\sigma$ an activation function, and $W^{(l)}$ the layer-specific trainable weights. We limit this study to GCN, as this is currently the only backbone implemented in BeGin~\citep{ko2024beginextensivebenchmarkscenarios}. Other popular backbones could be GraphSAGE and GAT, but a far-reaching extension of BeGin is outside the scope of this work. Additionally, GAT has many more parameters to learn, which would substantially increase calculation time. 

The main choices for the backbone are the depth and width of the layers~\citep{de2021continual, mirzadeh2022architecturematterscontinuallearning, Wei_2024}. The number of layers varies between 1 and 3. Here, a trade-off needs to be made. On the one hand, money laundering patterns often contain nodes that are a couple of hops in the network apart, requiring more layers to capture this information. On the other hand, having too many layers in a GNN leads to over-smoothing~\citep{Li_Han_Wu_2018}, lowering the predictive power of the model. 
The dimensions are the same for all hidden layers in the GCN. We test three values, namely 64, 128 and 256. 

Another choice concerns the number of epochs per task, which we set to 1, 2, 5 and 10. The more time a model is given to train on a single task, the more likely the weights are changed and previous knowledge is lost. On the other hand, the fewer epochs used, the harder it is for the model to learn and improve performance on the current task. 

Some hyperparameters of the GCN are fixed. 
For all experiments, we use the Adam optimizer with learning rate 0.001 and no weight decay and cross-entropy loss. The activation function for the GCN is ReLU, and the dropout rate is set to 0.5.

\subsection{Task Definition}
The elliptic data set contains fraud/non-fraud labels, and we therefore apply binary classification. 
We can take each time step as an individual task, resulting in 49 tasks. As this many tasks might pose a problem for the continual learning methods, we also group different time steps per task. Hence, we opt to also have seven time steps per task. The latter is chosen to have the same number of time steps in each task.

The patterns in the IBM data set each have their own separate label, resulting in multi-class classification.
For the IBM data set, the tasks are defined using the different patterns present in the data set. The first task will consist of two labels, i.e., legitimate transactions and a first money laundering pattern. Subsequent tasks are defined by adding one novel pattern at a time. We analyse the sensitivity to the order in which the tasks are presented. The different ways represent different scenarios that can occur when applying continual learning for AML in practice. 

A first way is to present the patterns in ascending order of difficulty. 
When a bank starts implementing AML procedures,  investigators are not yet experts in all patterns. They start by mastering simpler patterns, and gradually become more familiar with more complex money laundering patterns. This is also a result of the cat-and-mouse game between institutions and criminals. If financial institutions become better at detecting specific patterns, launderers will adapt and resort to more complex operations, prompting institutions to learn to detect these more complex patterns. To see if the complexity of the pattern plays a role, we also present the patterns in decreasing order of difficulty. 

The determination of what patterns are more difficult than others is inspired by the work of~\citet{egressy2024provably}. The authors extend message-passing GNNs step-by-step to prove that the extended GNNs can capture more patterns. We use those insights to order the patterns from least to most complex as follows: fan-in, fan-out, bipartite, gather-scatter, scatter-gather, stack, cycle and random. 

A second way to present the patterns is in descending order of frequency, i.e, gather-scatter, scatter-gather, stack, fan-out, fan-in, cycle, bipartite, random. It is hypothesized that more frequent patterns will be noticed sooner. This also makes the subsequent tasks more difficult to learn, since each time there are fewer observations to learn from. On the other hand, the model might be able to achieve better performance on these new tasks by transferring knowledge from previous tasks. In addition, we will also present the task in reverse order, from least to most frequent. 

Finally, as a baseline approach, we also present the patterns in random order, which for our experiments is fan-out, fan-in, gather-scatter, scatter-gather, cycle, random, bipartite and stack. 

Note that for the IBM data set, the transaction network is static. We incrementally learn the novel patterns, while keeping all nodes and edges the same. 

\subsection{Continual Learning Methods}
Different methods to prevent forgetting are present in continual graph learning literature, starting with the type of incremental learning to use. For the IBM data set, we use class incremental learning, since each new pattern is seen as its own class. The main problem as indicated in the literature is that these different tasks in the class-incremental setting are very similar, possibly resulting in strong forgetting across tasks~\citep{wang2024comprehensive, lee2021continual}. 

Domain-IL is used for the elliptic data set, where we recognise that the distribution in money laundering patterns can change. This mimics what happens in reality, since financial institutions will also use binary classification (i.e., legit vs. money laundering), but need to consider that the modus operandi of fraudsters evolves over time. 

The continual learning methods used are Gradient Episodic Memory (GEM)~\citep{GEM}, Elastic Weight Consolidation (EWC)~\citep{EWC}, Learning without forgetting (LWF)~\citep{lwf}, Memory Aware synapses (MAS)~\citep{mas}, and Topology-aware Weight Preserving (TWP)~\citep{twp}. This selection is made since the literature as presented in Section~\ref{subsec: CL fraud} also relies on these methods for fraud detection. 
These methods are compared to the bare and joint model. 
\begin{itemize}
    \item \textbf{Bare}: We iteratively fine-tune the model on only the data of the current task. In continual learning, this is taken as a lower bound for the performance. 
    \item \textbf{Gradient Episodic Memory (GEM)}~\citep{GEM}: GEM uses a fixed budget for memory allocation, and this memory is filled without any smart allocation of replay. The gradient of the current task is projected onto the space spanned by the gradients calculated using the replay examples, to avoid that the losses on previous tasks will increase. 
    \item \textbf{Elastic Weight Consolidation (EWC)}~\citep{EWC}: EWC uses the Fisher information matrix, based on the gradient of the loss, to find the weights that were important for the previous task. It changes the loss function by introducing heavier penalization of updating more important weights. 
    \item \textbf{Learning without forgetting (LWF)}~\citep{lwf}: LwF is introduced using a multi-task architecture, where part of the model is shared, and part is fine-tuned for each specific task. It assumes that no data of older tasks is available. LwF starts by constructing new `ground truth' labels by looking at the output on the parts of the old task using the data of the current task. Original capabilities are preserved by trying to keep these outputs as is while training on the new task. 
    \item \textbf{Memory Aware synapses (MAS)}~\citep{mas}: MAS uses the gradient of the squared $l2$-norm of the learned function output to express weight importance. Contrary to EWC, the weight importance determined by MAS is done in an unsupervised manner. 
    \item \textbf{Topology-aware Weight Preserving (TWP)}~\citep{twp}: TWP uses two sub-modules, one for task-related objectives and one for topology-related objectives. The task-related objective is similar to EWC where weight importance is measured via the gradient of the loss. The topology-related objective relies on the gradient vector of the attention coefficients in a GAT to incorporate network topology. The authors include a non-parametric proxy for the attention, in case the GNN backbone does not include an attention mechanism. 
    \item \textbf{Joint}: the joint model takes an accumulative approach. Similar to the bare model, the model of the previous task is fine-tuned on the current task. Contrary to the bare model, the joint model is fine-tuned using all data of the current and past tasks. In continual learning, this is taken as an upper bound for the performance. 
\end{itemize}

\subsection{Evaluation}
We focus on average forgetting and average performance. Previous work mostly uses accuracy to evaluate performance~\citep{https://doi.org/10.1002/widm.1526}. However, AML deals with strong class label imbalance, motivating the evaluation of performance by using the micro-F1 score for both data sets. 

In continual learning, special evaluation metrics are developed to reflect that the model is fine-tuned sequentially on the tasks. 
We use the model tuned for task $j$, and evaluate it on the test data of previous tasks $i\leq j$. This allows us to quantify the forgetting that has occurred after fine-tuning the model on task $j$.

Using this principle, we define the performance matrix~\citep{CGLB}, $M \in \mathcal{R}^{k\times k}$, with $k$ the number of tasks. The elements of the performance matrix are defined as:
\[ M_{i,j} = 
\begin{cases}
	\text{Performance on task } i \text{ after training on task } j & \text{if } i\leq j  \\
	0 & \text{otherwise}
\end{cases} 
\]
Hence, the performance matrix is a lower triangular matrix. The visualisation of the performance matrix using a heatmap is a first, qualitative evaluation of the methods. 

The performance matrix entries are used to calculate quantitative evaluation metrics, i.e., the average performance~(AP) and average forgetting~(AF). 
Average performance is the average micro-F1 over the tasks, after training on all tasks. Average forgetting compares the micro-F1 of a task after training on said tasks to the accuracy after learning on all tasks.
Using the performance matrix, we defined these metrics as~\citep{ko2024beginextensivebenchmarkscenarios}: 
\begin{eqnarray}
	\text{AP} & = & \sum_{i=1}^{k} \frac{M_{k,i}}{k} \\
	\text{AF} & = & \sum_{i=1}^{k-1}\frac{M_{i,i}-M_{k,i}}{k-1}
\end{eqnarray}

In the end, we are also interested in the performance of the final model on all tasks. Therefore, we include the micro-F1 score on all data after fine-tuning the model on the final task. 

\section{Results and Discussion}
\label{sec:resultsanddiscussion}
As shown in the pipeline of Figure~\ref{Figure3}, many hyperparameters choices are tested in this work. We start below with a general overview of all results in which some trends are already clear. Afterwards, a detailed discussion for each choice is given. 

We provide the full set of results on the average forgetting and average performance in the figures below. Figure~\ref{Figure5} contains a scatter plot of the results over all different (hyper-)parameters for the IBM data set. 
We see strong forgetting across experiments, which is probably caused by the strong similarity among the different tasks~\citep{wang2024comprehensive}. 

\begin{figure}
    \centering
    \includegraphics[width=0.8\linewidth]{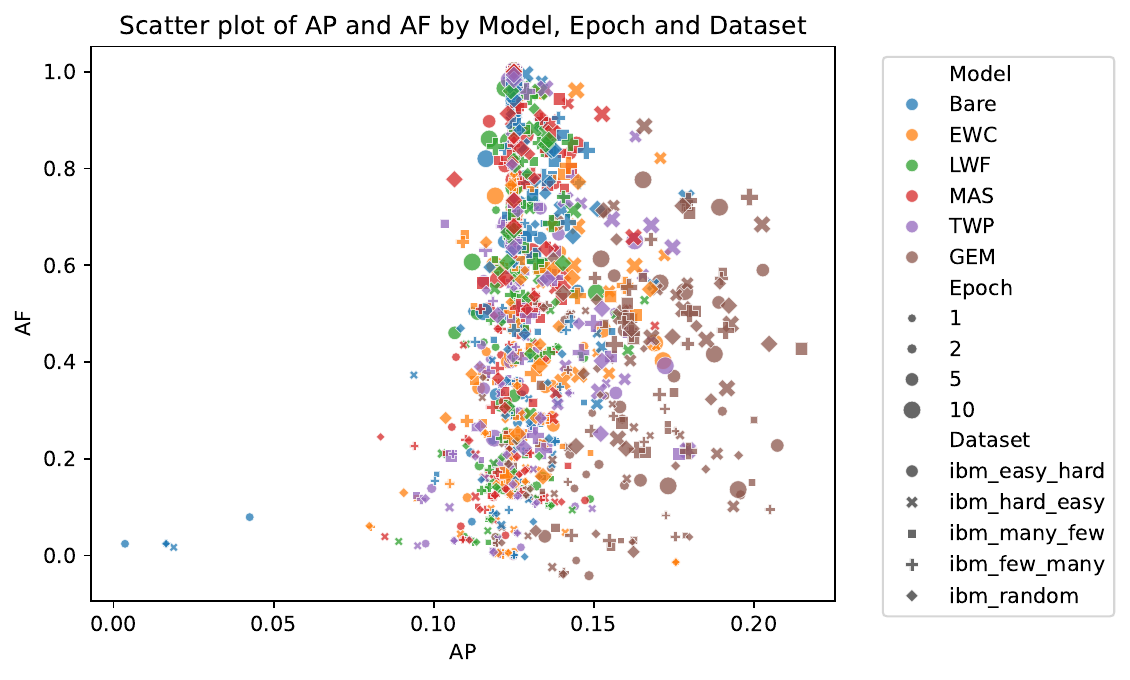}
    \caption{Scatterplot of the average forgetting plotted against the average performance for the IBM data set. }
    \label{Figure5}
\end{figure}

We notice that in general more epochs lead to more forgetting, while there is a limit on the performance the model can obtain. Additionally, GEM seems to achieve good performance without suffering too much forgetting. 

\begin{figure}
    \centering
    \includegraphics[width=0.8\linewidth]{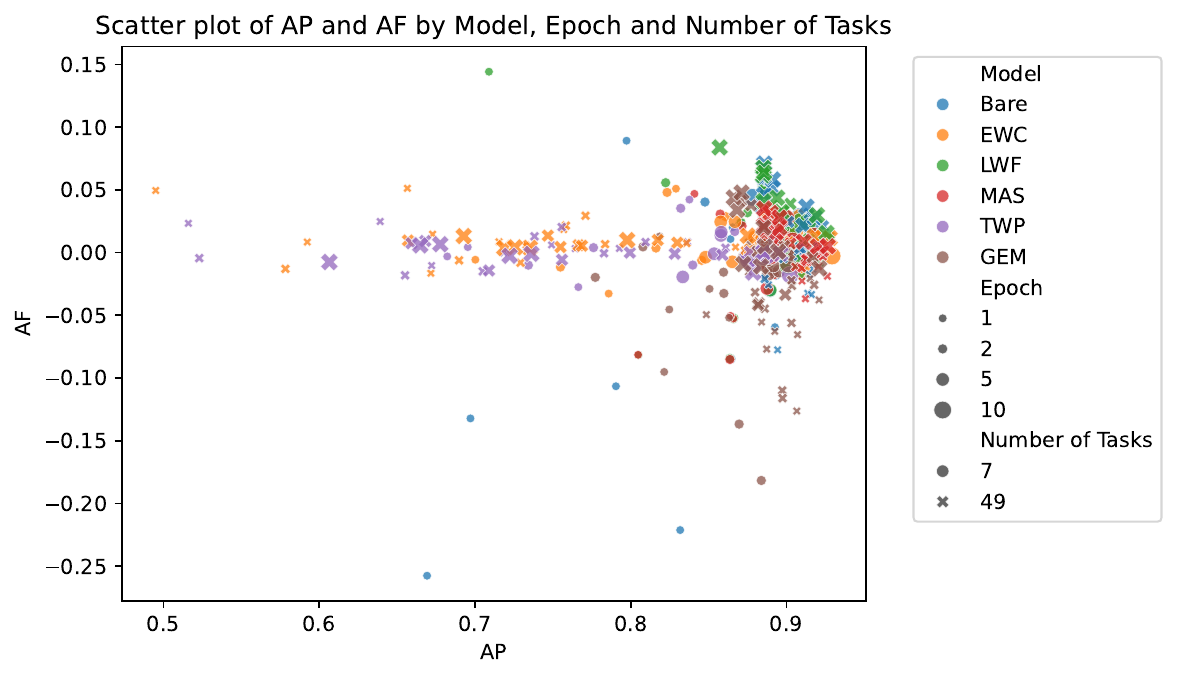}
    \caption{Scatterplot of the average forgetting plotted against the average performance for the Elliptic data set. }
    \label{Figure6}
\end{figure}

A similar figure is given for the Elliptic data set. Figure~\ref{Figure6} has different results. For the Elliptic data set, it seems that forgetting is less of a problem, but the average performance clearly increases with the number of epochs. In general, the results are similar between the setting with seven and the one with 49 tasks. The lack of forgetting can be because the distribution shift less severe in this data set. 

Looking at the methods, it seems that also here GEM has on average lowest forgetting, while LwF and Bare have higher forgetting. The difference in forgetting among methods for the Elliptic data set is, however, much less pronounced than for the IBM data set. 

\begin{figure}
    \centering
    \includegraphics[width=0.75\linewidth]{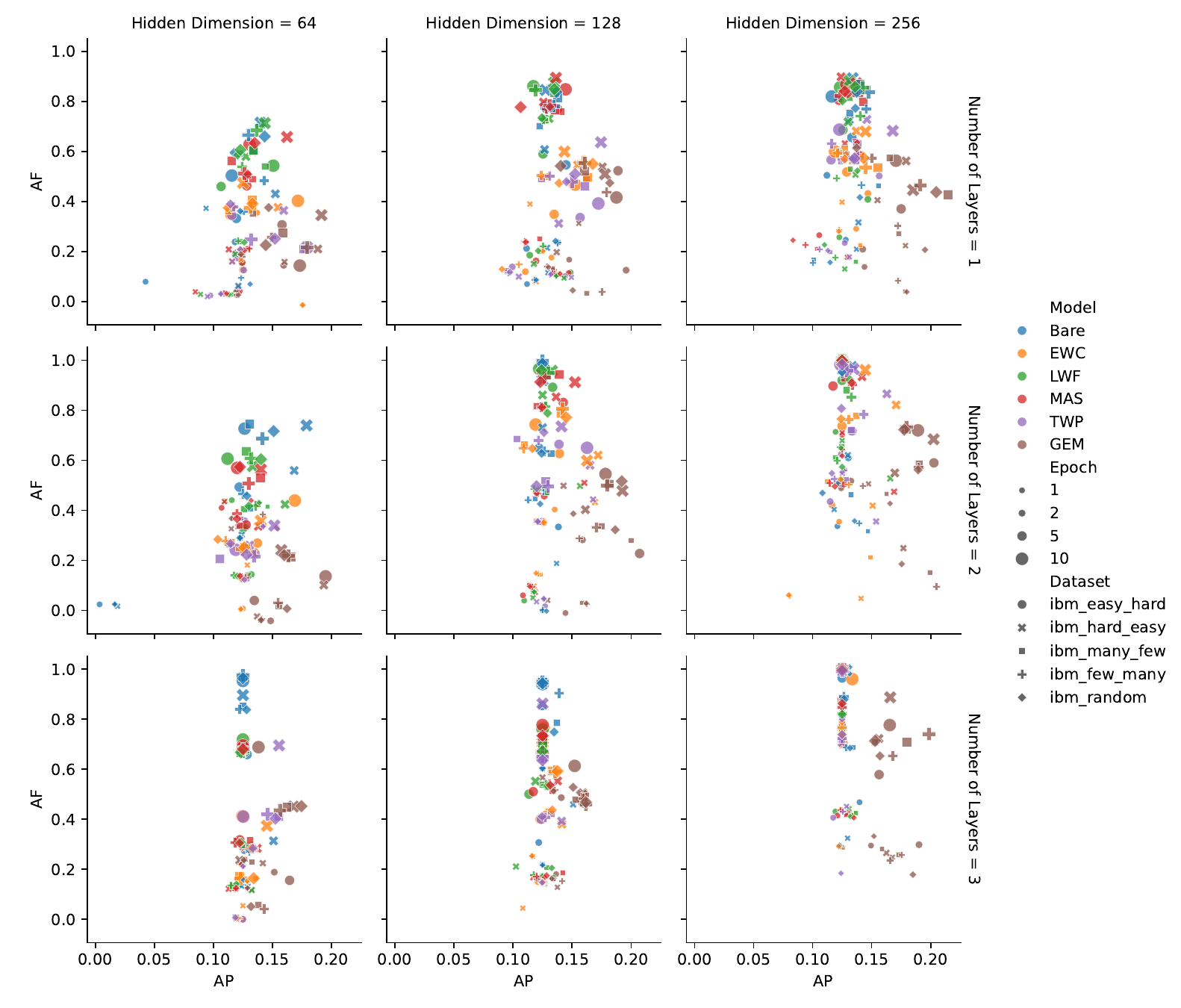}
    \caption{Scatter plots of the average forgetting (lower is better) plotted against the average performance (higher is better), split according to the breadth and width of the GCN.}
    \label{Figure7}
\end{figure}

A more fine-grained visualisation for the IBM data set is provided in Figure~\ref{Figure7}. The results are split in different plots according to the depth and width of the model. We see that higher dimensions lead to an increase in performance, but also correlate with stronger forgetting. We also note that the Bare model suffers strong forgetting, especially for the GCNs with three layers. For the other layers, Bare, MAS and LwF seem to consistently suffer more from forgetting for similar performance, compared to the other methods. 

\subsection{Number of Epochs}
\begin{figure}
    \centering
    \includegraphics[width=0.5\linewidth]{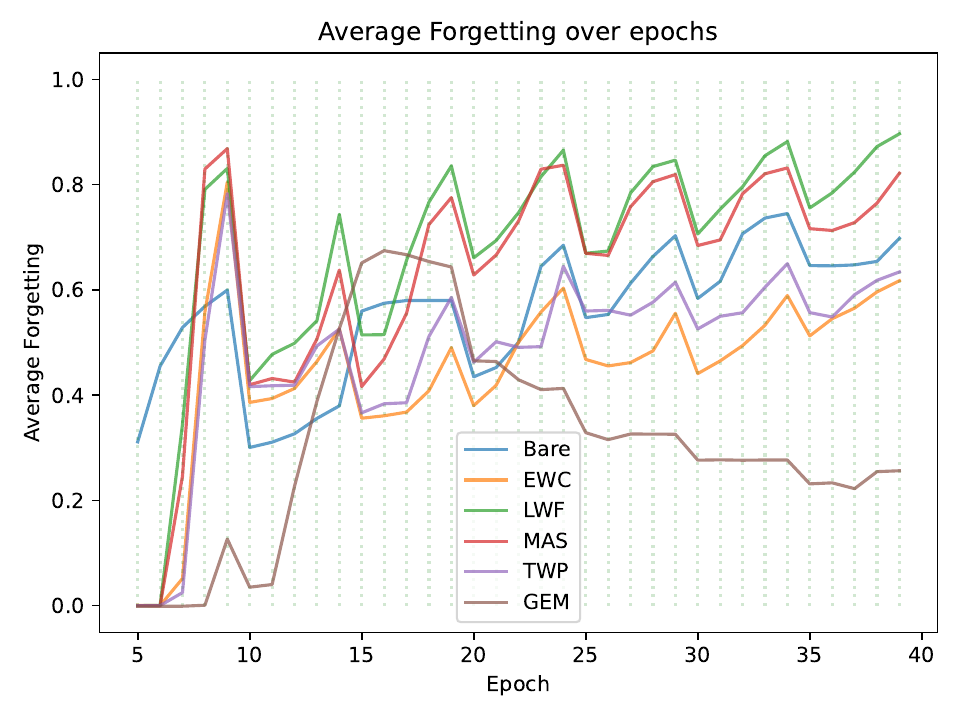}
    \caption{Average forgetting for the different methods on the IBM data set, where the backbone is a GCN with two layers, each having dimension 128, and we take five epochs per task.}
    \label{Figure8}
\end{figure}

To illustrate the importance of carefully choosing the number of epochs, the evolution of the average forgetting as a function of the number of epochs is shown in Figure~\ref{Figure8} for the IBM data set. We can clearly see that the average forgetting stays relatively low for a couple of epochs, but suddenly jumps up. This illustrates that, in this case, knowledge from previous tasks is not lost gradually, but suddenly. 

We note that the average forgetting drops when going to a new task, after every five epochs. This is because a new task is added to the average forgetting calculations. As the model has just fine-tuned on data from that task, the forgetting is expected to still be low. This results in a lower average, and hence a drop in forgetting when considering the next task. 

The average forgetting, average performance and final performance are reported for the IBM data set in Table~\ref{tab:results128 2 eh} over the different number of epochs per task. We see that performance improves when increasing the number of epochs. However, the amount of forgetting becomes a major issue with a higher number of epochs. It seems that only GEM is able to keep forgetting at a lower level. 

We see that the final performance goes down considerably when the number of epochs per task goes up. This is caused by the high forgetting for these cases. Especially forgetting for the first task is problematic, given that it contains the majority class. 

\begin{table}[h!]
\centering
\begin{tabular}{|c|c|c|c|c|c|c|}
\hline
Epochs & Bare & EWC & LWF & MAS & TWP & GEM \\ \hline
 & AP: 0.1250 & AP: 0.1353 & AP: 0.1095 & AP: 0.1084 & AP: 0.1272 & AP: \textbf{0.1444} \\ 
1 & AF: 0.0000 & AF: 0.4036 & AF: 0.0394 & AF: 0.0606 & AF: 0.0171 & AF: \textbf{-0.0100} \\ 
 & Fin.: 0.0004 & Fin.: 0.0003 & Fin.: \textbf{0.7800} & Fin.: 0.7431 & Fin.: 0.6529 & Fin.: 0.6372 \\ \hline
 & AP: 0.1385 & AP: 0.1259 & AP: 0.1253 & AP: 0.1265 & AP: 0.1250 & AP: \textbf{0.1588} \\ 
2 & AF: 0.3340 & AF: 0.3501 & AF: 0.4897 & AF: 0.4569 & AF: 0.3574 & AF: \textbf{0.2823} \\ 
 & Fin.: 0.0001 & Fin.: 0.0076 & Fin.: 0.0001 & Fin.: 0.0001 & Fin.: 0.0002 & Fin.: \textbf{0.0339} \\ \hline
 & AP: 0.1220 & AP: 0.1393 & AP: 0.1336 & AP: 0.1425 & AP: 0.1389 & AP: \textbf{0.2072} \\ 
5 & AF: 0.6489 & AF: 0.6274 & AF: 0.8924 & AF: 0.8318 & AF: 0.6642 & AF: \textbf{0.2277} \\ 
 & Fin.: 0.0001 & Fin.: 0.0002 & Fin.: 0.0001 & Fin.: 0.0001 & Fin.: 0.0002 & Fin.: \textbf{0.2783} \\ \hline
 & AP: 0.1250 & AP: 0.1192 & AP: 0.1222 & AP: 0.1250 & AP: 0.1627 & AP: \textbf{0.1783} \\ 
10 & AF: 0.9804 & AF: 0.7433 & AF: 0.9657 & AF: 0.9576 & AF: 0.6503 & AF: \textbf{0.5456} \\ 
 & Fin.: 0.0001 & Fin.: 0.0001 & Fin.: 0.0001 & Fin.: 0.0001 & Fin.: 0.0003 & Fin.: \textbf{0.2337} \\ \hline
\end{tabular}
\caption{Performance metrics (AP and AF) and final performance on all data (Fin.) for different models and epochs for the IBM data set. The backbone is a GCN with two layers, each having dimension 128.}
\label{tab:results128 2 eh}
\end{table}

The average forgetting, average performance and final performance on all data are reported for the Elliptic data set in Table~\ref{tab:results128 2 elliptic 7} for seven tasks and in Table~\ref{tab:results128 2 elliptic 49} for 49 tasks over the different number of epochs per task. As before, the performance increases with the number of epochs per task, while forgetting increases, but not as drastically as for the IBM data set. Here, both TWP and GEM seem to be able to achieve high performance in combination with negative forgetting, meaning that the model also improves its performance on previous tasks when fine-tuning on novel tasks. 

The final performance of the models seems less affected by the number of epochs. This is probably caused by the fact that there is very little forgetting since there is limited data shift over the tasks. Therefore, the model has had ample training time at the end of fine-tuning on all tasks, even when the number of epochs per task is small. 

\begin{table}[h!]
\centering
\begin{tabular}{|c|c|c|c|c|c|c|}
\hline
Epochs & Bare & EWC & LWF & MAS & TWP & GEM \\ \hline
 & AP: 0.8318 & AP: 0.8903 & AP: 0.8908 & AP: \textbf{0.8922} & AP: 0.8900 & AP: 0.8860 \\ 
1 & AF: \textbf{-0.2213} & AF: -0.0010 & AF: 0.0012 & AF: 0.0000 & AF: -0.0018 & AF: -0.0196 \\ 
 & Fin.: 0.8284 & Fin.: 0.9019 & Fin.: 0.9029 & Fin.: \textbf{0.9041} & Fin.: 0.9020 & Fin.: 0.8930 \\ \hline
 & AP: 0.8971 & AP: 0.8936 & AP: 0.8946 & AP: \textbf{0.8973} & AP: 0.8936 & AP: 0.8696 \\ 
2 & AF: -0.0012 & AF: -0.0035 & AF: 0.0012 & AF: -0.0026 & AF: -0.0011 & AF: \textbf{-0.1368} \\ 
 & Fin.: 0.9074 & Fin.: 0.9028 & Fin.: 0.9057 & Fin.: \textbf{0.9077} & Fin.: 0.9050 & Fin.: 0.8771 \\ \hline
 & AP: 0.9004 & AP: 0.9037 & AP: 0.9040 & AP: 0.8978 & AP: 0.9063 & AP: \textbf{0.9097} \\ 
5 & AF: 0.0156 & AF: 0.0045 & AF: 0.0158 & AF: 0.0166 & AF: -0.0093 & AF:\textbf{ -0.0107} \\ 
 & Fin.: 0.9102 & Fin.: 0.9113 & Fin.: 0.9133 & Fin.: 0.9085 & Fin.: 0.9150 & Fin.: \textbf{0.9168} \\ \hline
 & AP: 0.9084 & AP: 0.9156 & AP: 0.9091 & AP: 0.8996 & AP: 0.9162 & AP: \textbf{0.9175} \\ 
10 & AF: 0.0287 & AF: 0.0116 & AF: 0.0248 & AF: 0.0275 & AF: \textbf{-0.0026} & AF: -0.0023 \\ 
 & Fin.: 0.9149 & Fin.: 0.9204 & Fin.: 0.9157 & Fin.: 0.9099 & Fin.: \textbf{0.9229} & Fin.: 0.9227 \\ \hline
\end{tabular}
\caption{Performance metrics (AP and AF) and final performance on all data (Fin.) for different models and epochs for the Elliptic data set, with seven tasks. The backbone is a GCN with two layers, each having dimension 128.}
\label{tab:results128 2 elliptic 7}
\end{table}

\begin{table}[h!]
\centering
\begin{tabular}{|c|c|c|c|c|c|c|}
\hline
Epochs & Bare & EWC & LWF & MAS & TWP & GEM \\ \hline
 & AP: \textbf{0.9146} & AP: 0.7328 & AP: 0.8978 & AP: 0.9129 & AP: 0.7927 & AP: 0.8756 \\ 
1 & AF: \textbf{-0.0173} & AF: 0.0060 & AF: 0.0105 & AF: -0.0081 & AF: 0.0033 & AF: -0.0155 \\ 
 & Fin.:\textbf{ 0.9229} & Fin.: 0.7200 & Fin.: 0.9100 & Fin.: 0.9213 & Fin.: 0.7821 & Fin.: 0.8778 \\ \hline
 & AP: 0.9144 & AP: 0.7584 & AP: 0.9157 & AP: \textbf{0.9203} & AP: 0.7557 & AP: 0.8798 \\ 
2 & AF: 0.0023 & AF: 0.0215 & AF: -0.0014 & AF: -0.0107 & AF: 0.0204 & AF:\textbf{ -0.0316} \\ 
 & Fin.: 0.9236 & Fin.: 0.7431 & Fin.: 0.9250 & Fin.: \textbf{0.9269} & Fin.: 0.7472 & Fin.: 0.8824 \\ \hline
 & AP: 0.8870 & AP: 0.8764 & AP: 0.8913 & AP: 0.9133 & AP: 0.8766 & AP: \textbf{0.9168} \\ 
5 & AF: 0.0507 & AF: 0.0024 & AF: 0.0372 & AF: \textbf{-0.0058} & AF: 0.0005 & AF: -0.0035 \\ 
 & Fin.: 0.9031 & Fin.: 0.8707 & Fin.: 0.9058 & Fin.: \textbf{0.9225} & Fin.: 0.8860 & Fin.: 0.9211 \\ \hline
 & AP: 0.8913 & AP: 0.7977 & AP: 0.8849 & AP: 0.8897 & AP: 0.7362 & AP: \textbf{0.8947} \\ 
10 & AF: 0.0583 & AF: 0.0100 & AF: 0.0617 & AF: 0.0316 & AF: \textbf{-0.0010} & AF: 0.0009 \\ 
 & Fin.: \textbf{0.9050} & Fin.: 0.7794 & Fin.: 0.9020 & Fin.: 0.9044 & Fin.: 0.7182 & Fin.: 0.8961 \\ \hline
\end{tabular}
\caption{Performance metrics (AP and AF) and final performance on all data (Fin.) for different models and epochs for the Elliptic data set, with 49 tasks. The backbone is a GCN with two layers, each having dimension 128.}
\label{tab:results128 2 elliptic 49}
\end{table}

\begin{figure}
    \centering
    \includegraphics[width=0.7\linewidth]{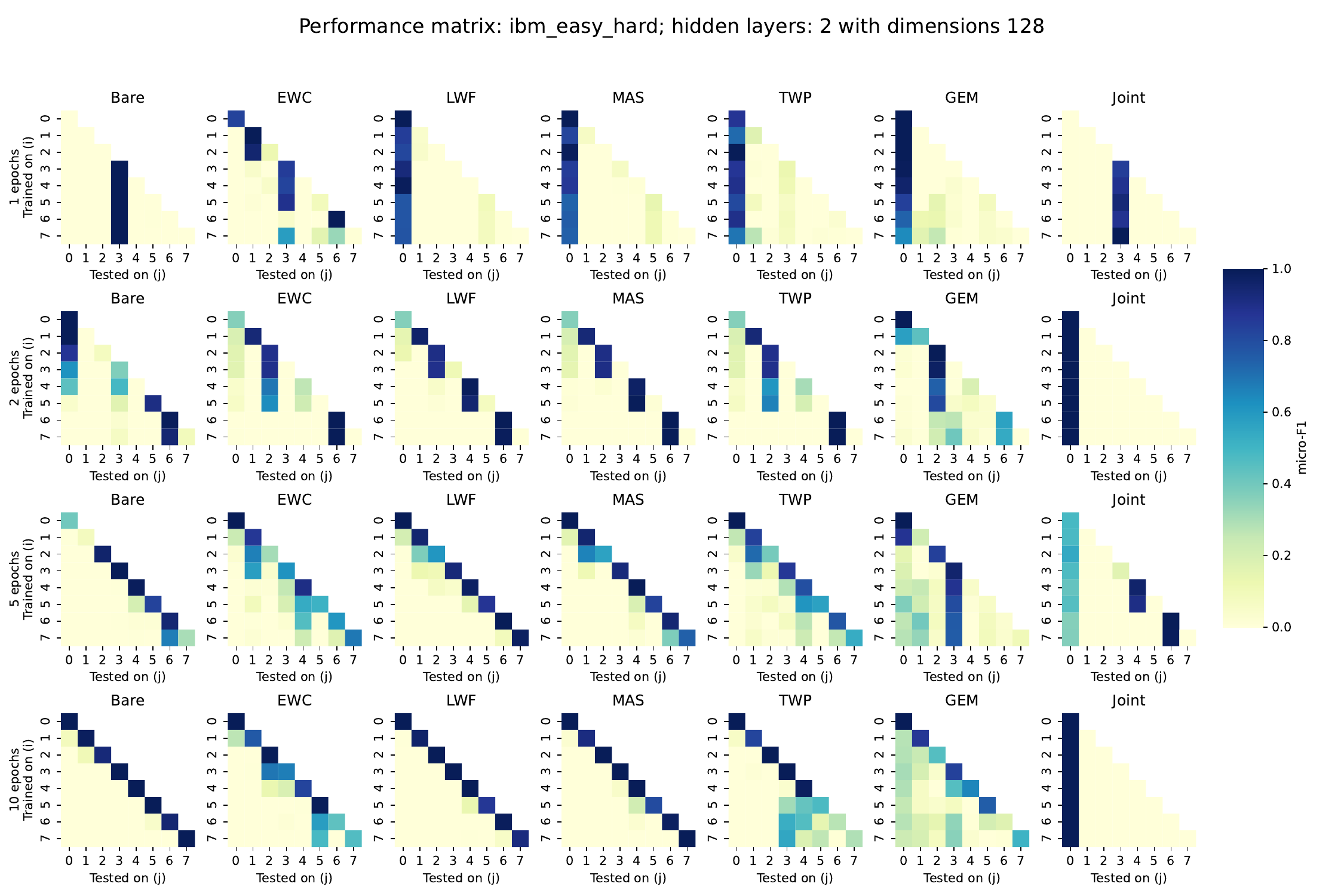}
    \caption{The performance matrices for the different methods for a varying number of epochs for the IBM data set.}
    \label{Figure9}
\end{figure}

Something we notice for the IBM data set in Figure~\ref{Figure8} and Table~\ref{tab:results128 2 eh} is that the Bare model, although considered as a lower bound, does not suffer the strongest forgetting of all models. There can be a couple of reasons for this. A first reason is that the network structure, via the inter-task connections, is beneficial for the bare model to retain knowledge from previous tasks. This reason seems to be supported by the results of the Elliptic data set in Table~\ref{tab:results128 2 elliptic 7}, where the Bare model seems to be performing worse, although not in all cases. Inter-task connections are absent in the Elliptic data set. 
A second reason for the deviant performance of the Bare model on the IBM data set is visible in the performance matrices in Figure~\ref{Figure9}. It seems that for some tasks, the bare model has difficulty obtaining good performance, leading to less \textit{learned information} to forget. 

\begin{figure}
    \centering
    \includegraphics[width=0.7\linewidth]{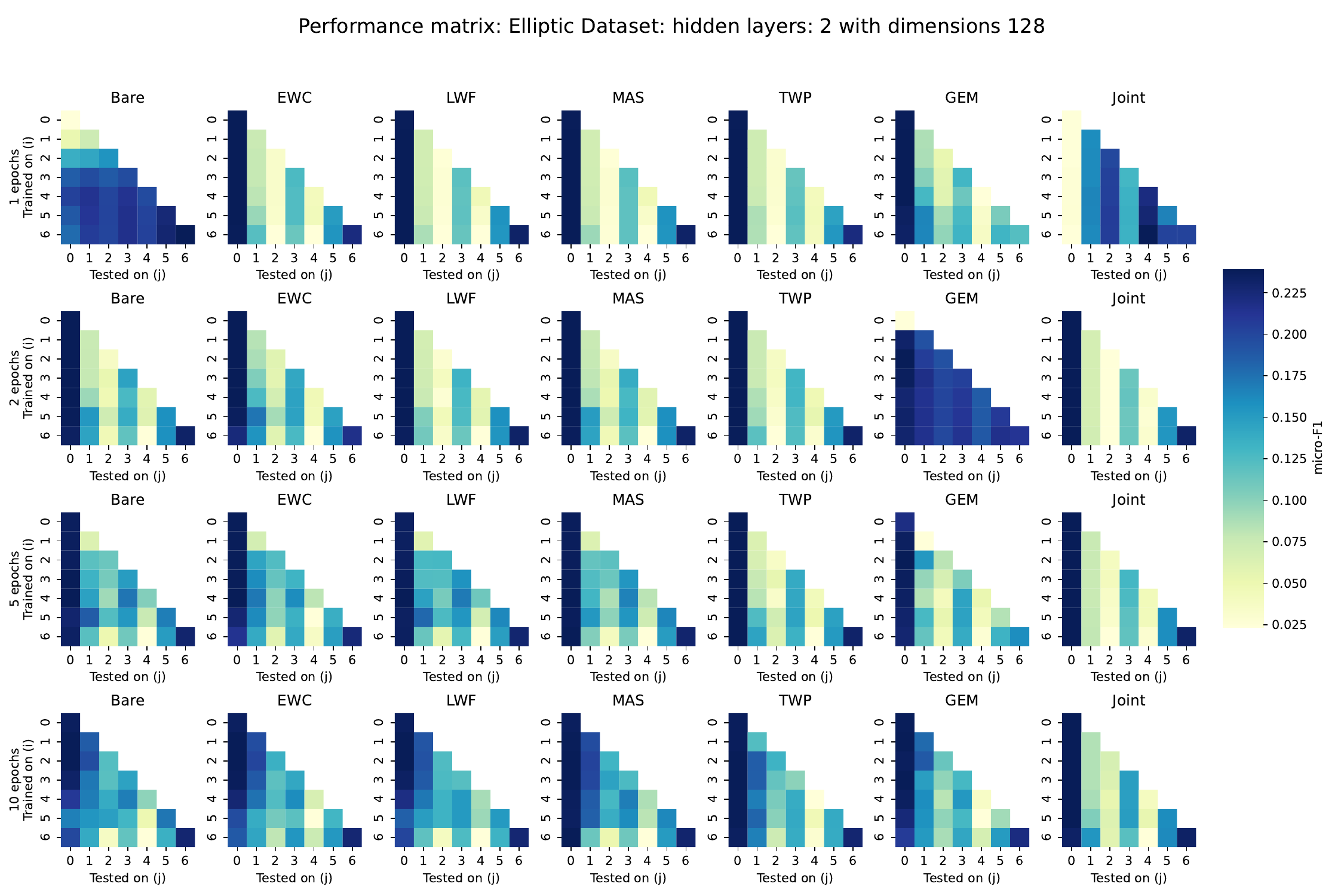}
    \caption{The performance matrices for the different methods for a varying number of epochs for the Elliptic data set with 7 time steps per task.}
    \label{Figure10}
\end{figure}

\begin{figure}
    \centering
    \includegraphics[width=0.7\linewidth]{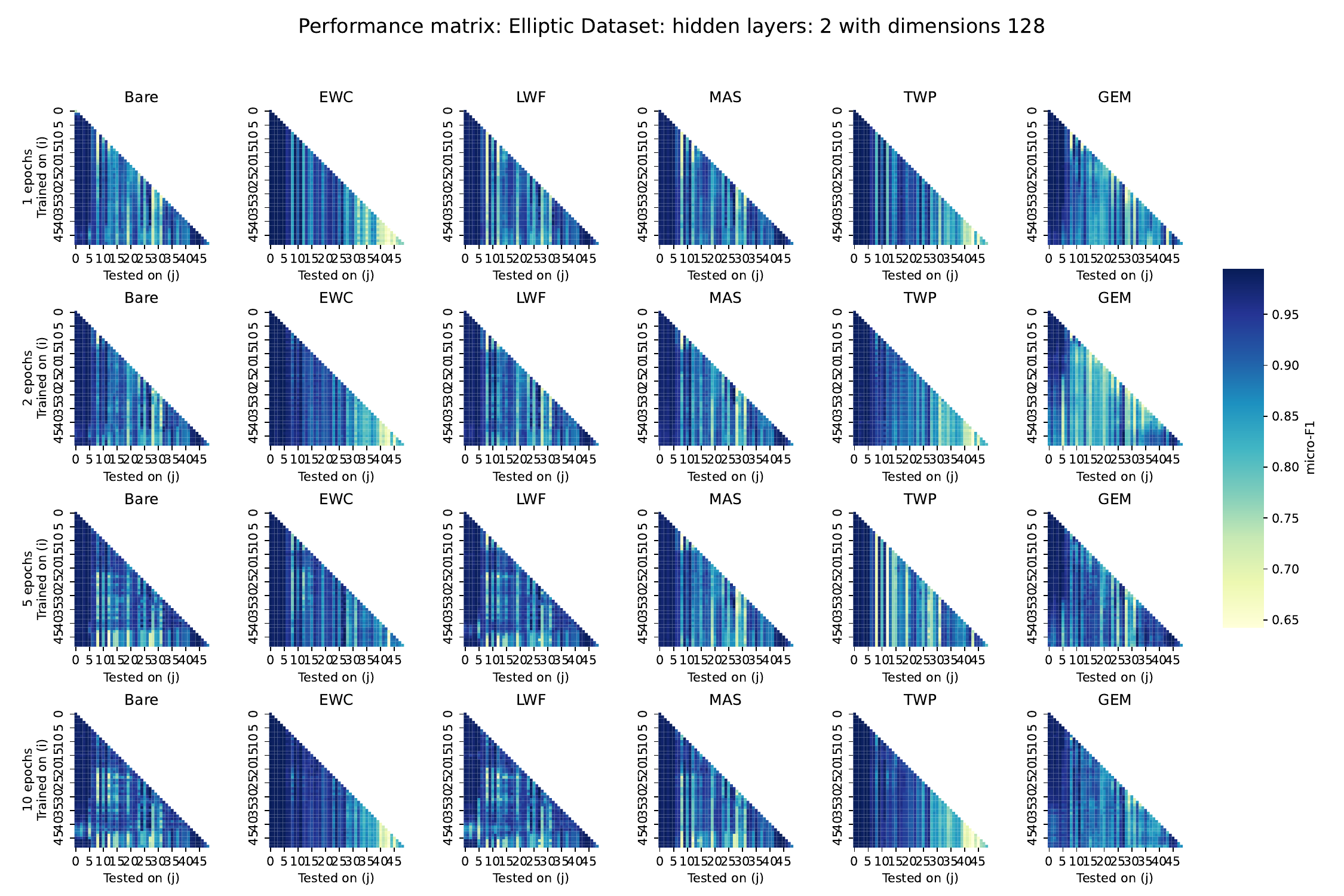}
    \caption{The performance matrices for the different methods for a varying number of epochs for the Elliptic data set with each time step a separate task.}
    \label{Figure11}
\end{figure}

Further analysis of the performance matrices in Figures~\ref{Figure9}-\ref{Figure11} illustrate that a balance needs to be struck between performance and forgetting. The forgetting is kept low with a lower number of epochs, although the model does not seem to learn any knowledge from the new tasks. On the other hand, the models tend to overfit on the latest task for the IBM data set if epochs are set too high. 

The complete collection of performance matrices is available as supplementary material online on Github\footnote{\url{https://github.com/VerbekeLab}}.

\subsection{Order of Patterns}
\begin{figure}
    \centering
    \includegraphics[width=0.8\linewidth]{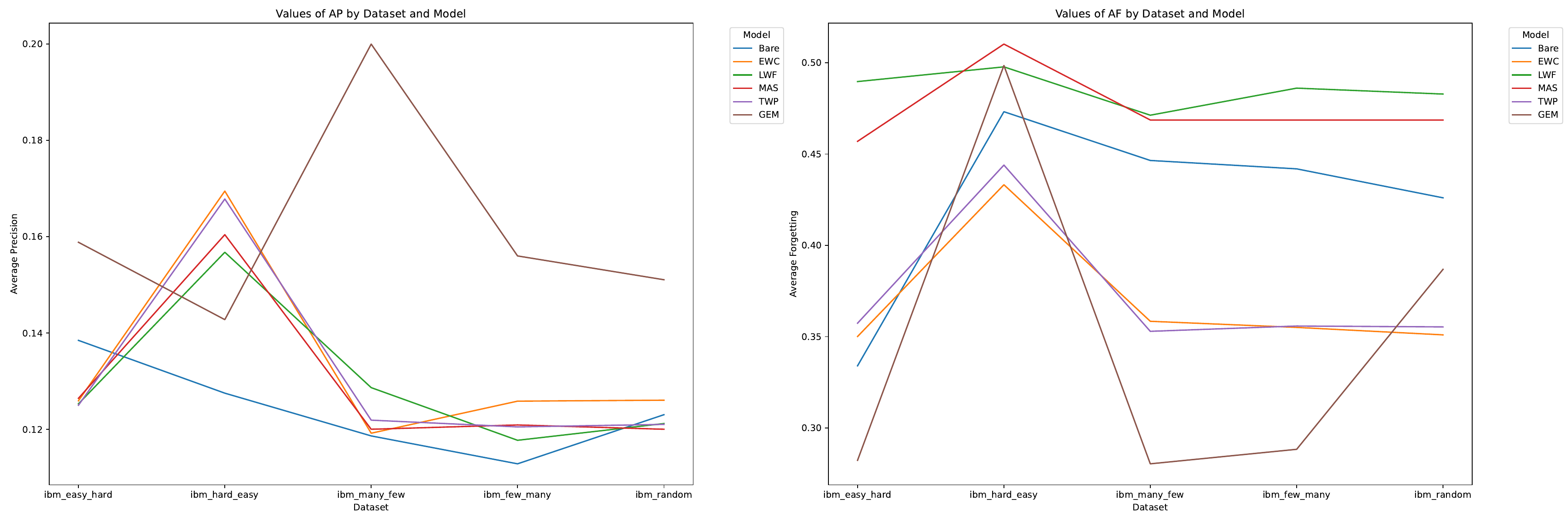}
    \caption{Global average of the average performance (left) and average forgetting (right) for the methods across the different permutations of the patterns.}
    \label{Figure12}
\end{figure}

The analysis of the order of patterns is only relevant for the IBM data set.
For the same architecture as before, we see in Figure~\ref{Figure12} that the forgetting and precision is more or less stable across different pattern orders. Only for \textit{hard to easy} we notice that the forgetting spikes for a couple of methods. Here, the average performance is also higher. 

\begin{figure}
    \centering
    \includegraphics[width=0.8\linewidth]{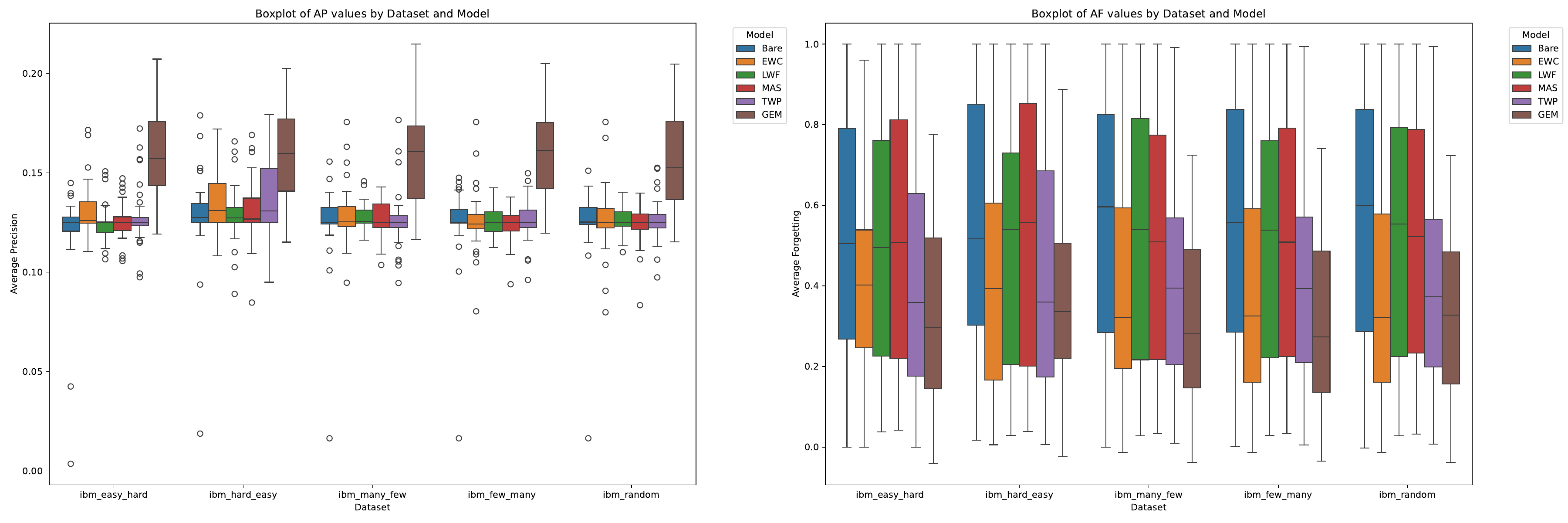}
    \caption{Boxplots of the average performance (left) and average forgetting (right) for the methods across the different permutations of the patterns.}
    \label{Figure13}
\end{figure}

When aggregating on all experiments, as shown in Figure~\ref{Figure13}, we see that on average the order of the patterns does not seem to have a major impact on the performance nor on the forgetting. This is in line with previous studies~\citep{de2021continual,nguyen2019understandingcatastrophicforgettingcontinual}.

We notice that the boxes for EWC and TWP - which is based on EWC - go a bit higher in terms of performance in the hard-to-easy setting, than for the others. This might indicate that these regularisation-based methods perform a bit better in this setting, although differences are minor. 

\subsection{Architecture of Backbone}
\begin{figure}
    \centering
    \includegraphics[width=0.8\linewidth]{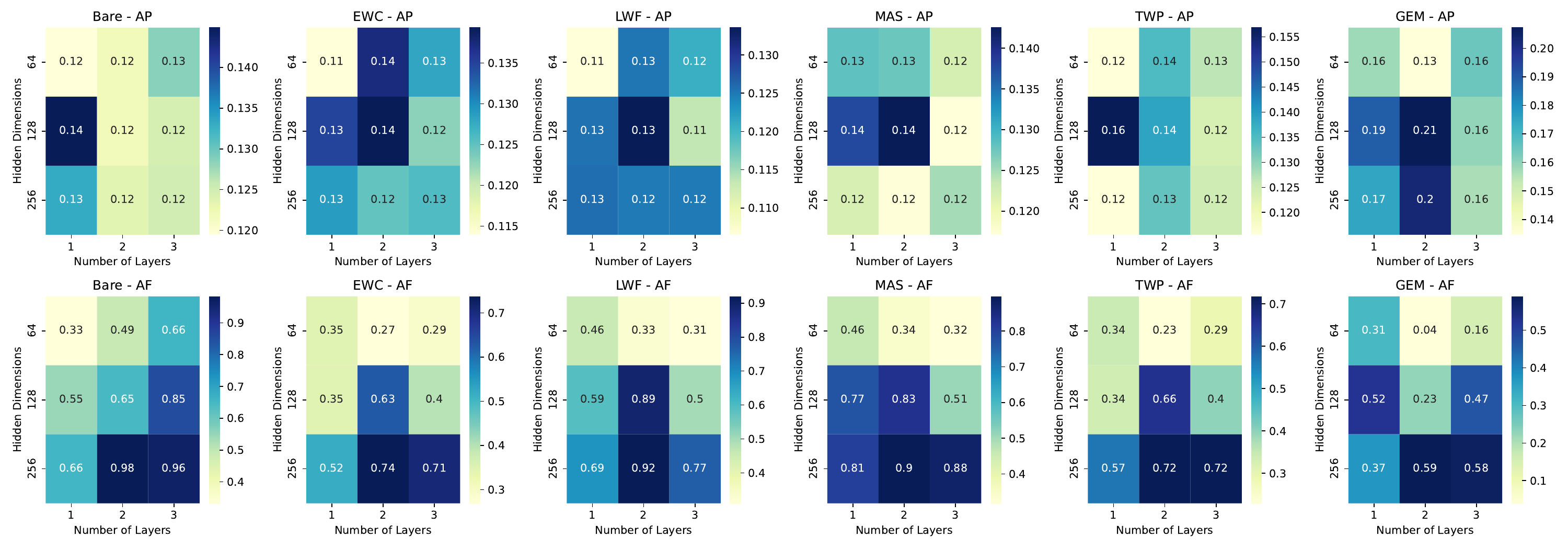}
    \caption{The average performance and average forgetting across different depths and widths of the GCN, when training for five epochs per task on the IBM data set.}
    \label{Figure14}
\end{figure}

\begin{figure}
    \centering
    \includegraphics[width=0.8\linewidth]{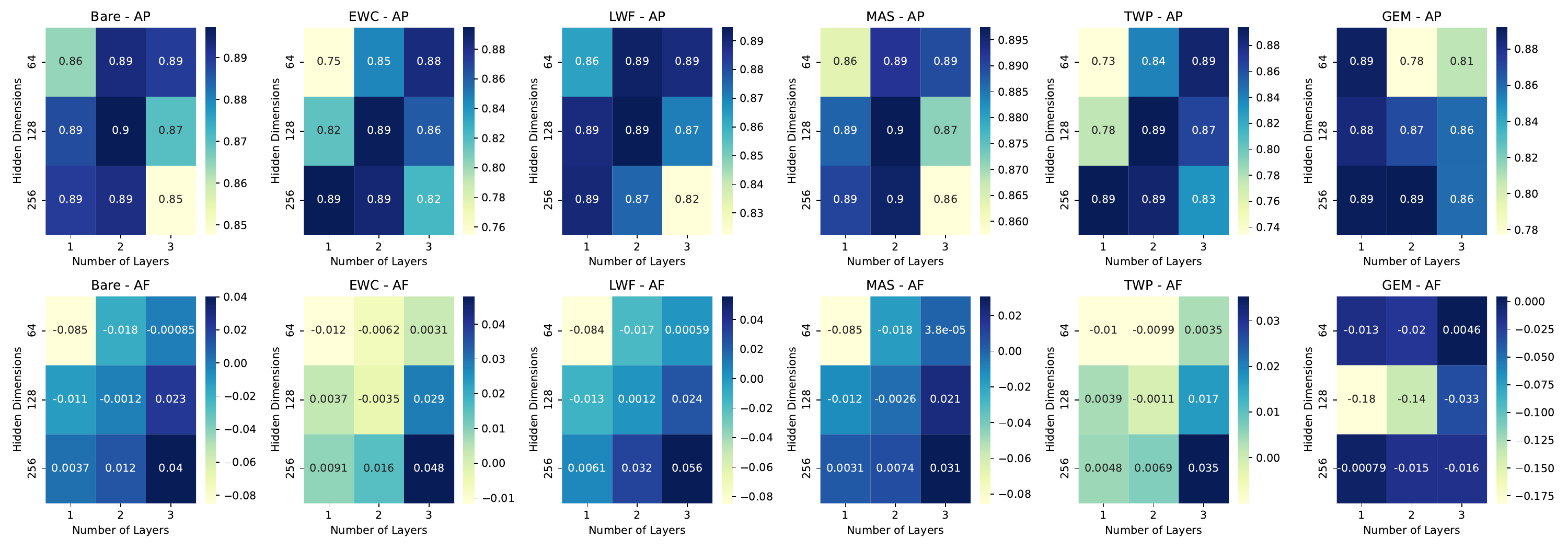}
    \caption{The average performance and average forgetting across different depths and widths of the GCN, when training for five epochs per task on the elliptic data set with seven tasks.}
    \label{Figure15}
\end{figure}

\begin{figure}
    \centering
    \includegraphics[width=0.8\linewidth]{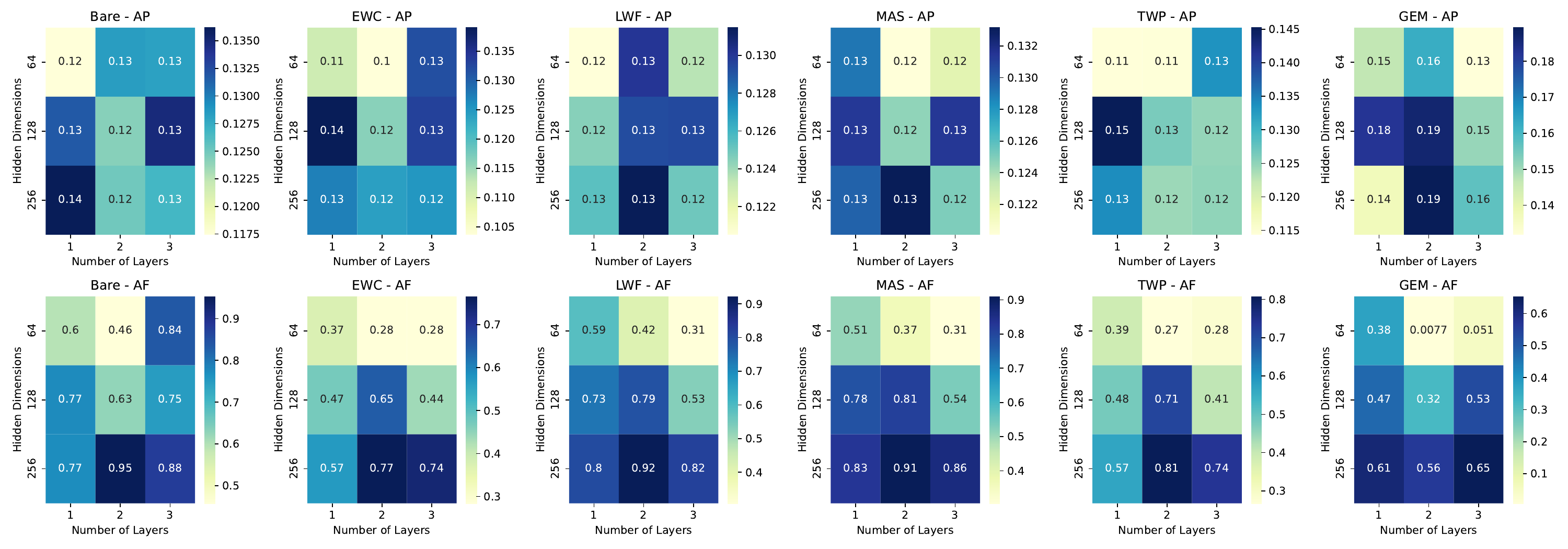}
    \caption{The average performance and average forgetting across different depths and widths of the GCN, when training for five epochs per task on the elliptic data set with 49 tasks.}
    \label{Figure16}
\end{figure}

In this section, we set out to answer \ref{RQ DW} by analysing the effect of the depth and width of the GCN model. 
The results in Figure~\ref{Figure7} already show that the architecture of the GCN - both in terms of depth and width - has an impact on performance and forgetting. We provide heatmaps of the average performance and forgetting for all methods to have a detailed view on the results. The results for the IBM data set on the easy-to-hard data set for five epochs in Figure~\ref{Figure14}. The results for the Elliptic data set with seven and 49 tasks for five epochs in Figure~\ref{Figure15} and Figure~\ref{Figure16}, respectively. 

Although intuitively more layers should be better in terms of average performance, the results do not show a clear dominance of two or three layers over one layer in the GCN. On the other hand, we can clearly see that forgetting is more severe if the GCN has more parameters. Deeper and wider GCNs tend to overfit more on the latest task. 

\subsection{Continual Learning Method}
To evaluate the effect of the continual learning method to answer \ref{RQ method}, we first look at the performance matrices in Figure~\ref{Figure8}. Visually, it seems that GEM most often retains previous knowledge both for a low as well as high number of epochs, while also learning the new task. 

\begin{figure}
    \centering
    \includegraphics[width=0.9\linewidth]{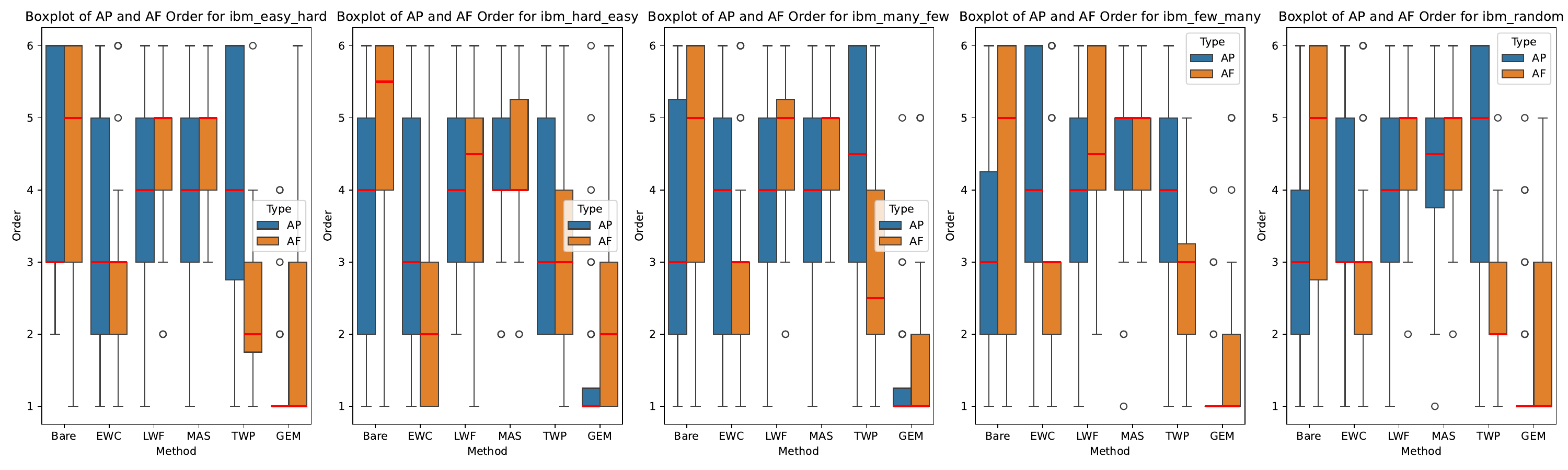}
    \caption{Box plot of the order of the different methods per permutation of the patterns on the IBM data set.}
    \label{Figure17}
\end{figure}

\begin{figure}
    \centering
    \includegraphics[width=0.9\linewidth]{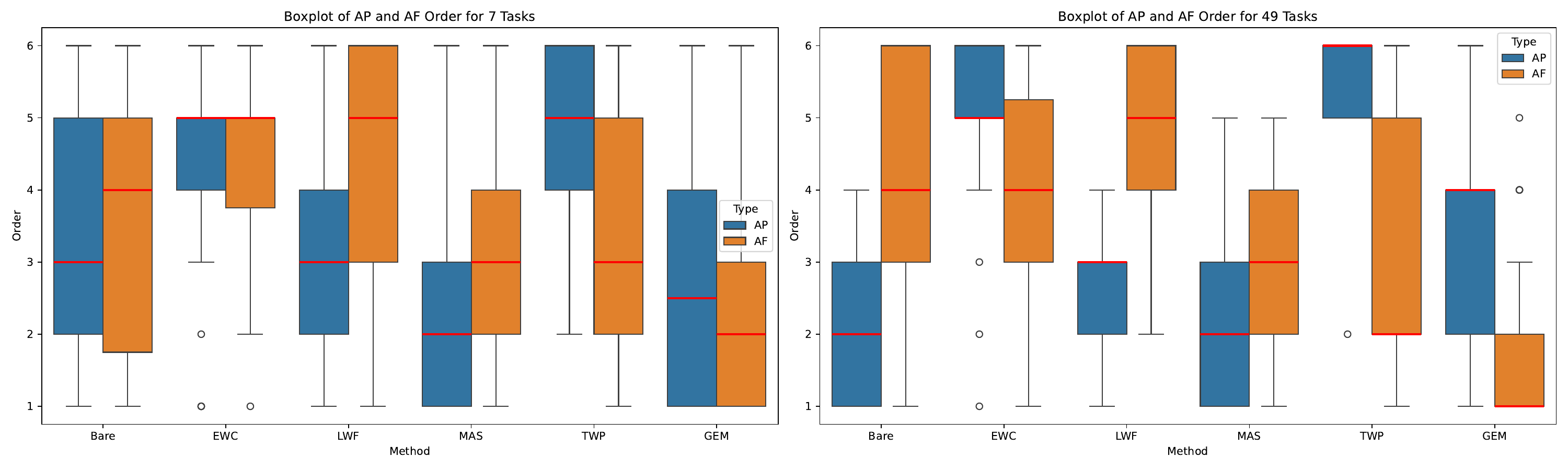}
    \caption{Box plot of the order of the different methods for the different number of tasks on the elliptic data set.}
    \label{Figure18}
\end{figure}

We confirm this by assessing the box plots of the ranking of the models in Figure~\ref{Figure17} and Figure~\ref{Figure18} for the IBM and Elliptic data set, respectively. On average, GEM scores best both in terms of low average forgetting as well as high average performance for both data sets. 

When looking at the IBM results, EWC and to a lesser extent TWP seems to also perform quite strongly. For the other methods, it seems that there is a trade-off between forgetting and performance. 

For the Elliptic data set, the results are slightly different, here MAS performs well, while EWC is performing quite poorly. For the other methods, we again see a trade-off between forgetting and performance. 

An observation made for all iterations is that continual learning methods seem to result in better AML detection methods than the bare and joint model. Looking at the performance matrices of the joint model, we see strong performance for the first task, concerning the majority class, but low performance on the other tasks. This results from the extreme class imbalance in AML. The joint GCN model seems to learn to always predict the majority class when minimising the loss function. 

\begin{figure}
    \centering
    \includegraphics[width=0.9\linewidth]{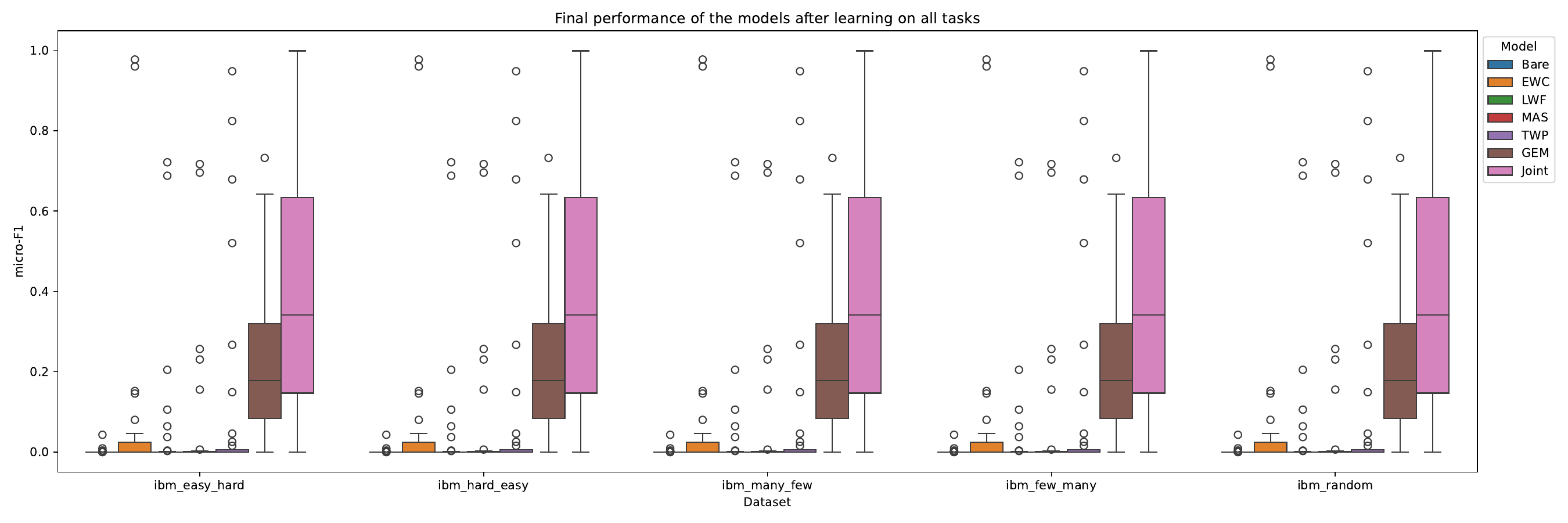}
    \caption{Box plot of the performance of the final model on all test data for the IBM data set.}
    \label{Figure19}
\end{figure}

We also consider the performance of the model on the full test set, after seeing all tasks. The results on the IBM data set are given in Figure~\ref{Figure19}. As expected, the joint model outperforms all other models, since it was trained on all data. We also see that GEM performs better than the other continual learning and bare models. This is due to the combination of high average performance and low average forgetting. 

\begin{figure}
    \centering
    \includegraphics[width=0.9\linewidth]{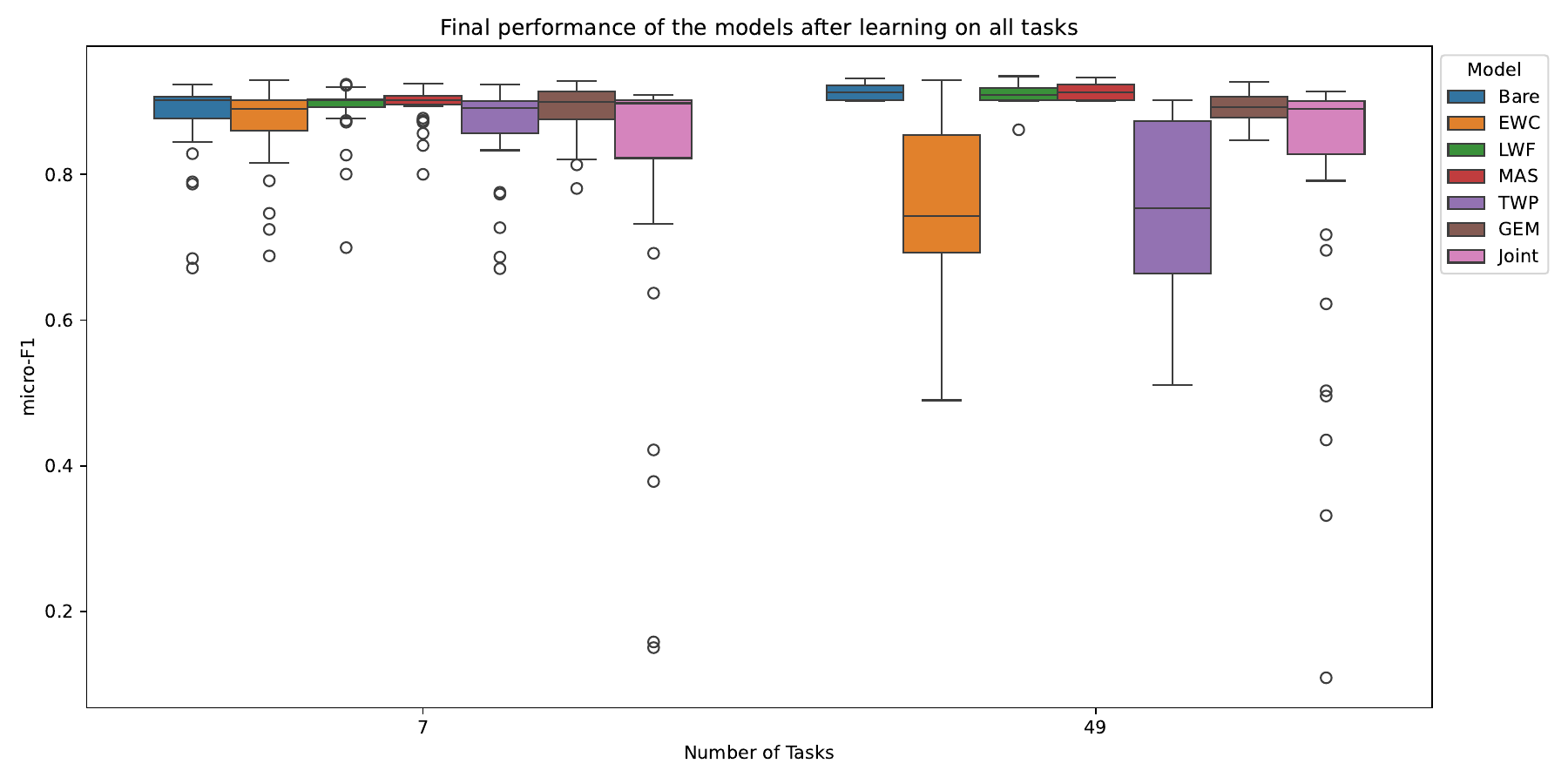}
    \caption{Boxplot of the performance of the final model on all test data for the IBM data set.}
    \label{Figure20}
\end{figure}

Similar results are provided for the Elliptic data set in Figure~\ref{Figure20}. Here, the picture is less pronounced that before. We see consistently strong results for Bare, LwF, MAS and GEM. Surprisingly, the joint model seems to perform slightly worse than these models. Given that it is trained on all data simultaneously, the joint model might suffer from the inclusion of the sudden closure of a dark market at time step 43~\citep{weber2019anti}.

\section{Conclusion}
\label{sec:conclusion}
Continual learning is essential in AML because (1) millions of transactions need to be monitored continuously, (2) fraud tactics are constantly evolving, causing underlying data distributions to shift, and (3) regulatory constraints often limit the amount of historical data that can be stored. Therefore, this work addresses four key research questions.

We started with reviewing the current state of the continual graph learning literature for AML (\ref{RQ lit}). We conclude that despite its ability to tackle these challenges, continual learning for AML has received limited interest in the scientific literature.

To expand on the current body of knowledge, we present the results of a comprehensive experiment on continual graph learning on two AML data sets. We presented experiments for node and edge classification. We investigated the effect of the hyperparameters including the task order (\ref{RQ hyp}), the effect of the GNN architecture (\ref{RQ DW}), and the different continual learning methods (\ref{RQ method}). 

We conclude that increasing the number op epochs per task too much may lead to overfitting on the present task, and hence to forgetting. 
With regard to the task order, our experiments are in line with previous work and confirm that there is no significant effect of the task order on performance. 

Based on the experimental results, we conclude that wide models are more prone to forgetting, and a balance needs to be struck between capturing longer money laundering chains and avoiding over-smoothing and forgetting when setting the depth of the GNN. 

Across the experiments, GEM performed well with minimal forgetting. This indicates that replay methods are best when it comes to AML. As noted, their application in practice might be hindered by regulations limiting the storage of transaction data. 

A surprising result is obtained regarding the joint model. The experiments show that continual learning methods are better at allowing the model to learn the different fraud patterns. When provided with all data, the joint model learns to consistently predict the majority class. 

Based on the presented literature review and experimental evaluation, we identify a series of directions for future work. First of all, a deeper analysis is needed on the specific challenges in fraud detection and how these can be addressed by continual learning methods. Future research should investigate the effect of rotating between patterns, and the effect of having periods with no fraud cases, in a continual learning setting.

Second, the BeGin framework should be extended to incorporate domain-incremental learning for edge classification, needed for extended analysis of the IBM data set, and to include more backbone architectures. This would result in extended experiments to complement the analysis done in this work. 

Finally, some of the problems present for AML are also present in other domains, especially the high class imbalance. Qualitative and quantitative research on the interplay between the degree of imbalance and the performance of continual learning is still lacking. 

\section*{Author Contributions}
\textbf{Bruno Deprez:} Conceptualization, Data Curation, Methodology, Software, Visualization, Writing – Original Draft Preparation.
\textbf{Wei Wei:} Conceptualization, Methodology, Software, Writing – Review \& Editing.
\textbf{Wouter Verbeke:} Conceptualization, Supervision, Writing – Review \& Editing.
\textbf{Bart Baesens:} Conceptualization, Supervision, Writing – Review \& Editing.
\textbf{Kevin Mets:} Conceptualization, Methodology, Supervision, Writing – Review \& Editing.
\textbf{Tim Verdonck:} Conceptualization, Methodology, Supervision, Writing – Review \& Editing.

\section*{Declaration of Competing Interest}
The authors declare that they have no known competing financial interests or personal relationships that could have appeared to influence the work reported in this paper.

\section*{Acknowledgements}
This work was supported by the Research Foundation – Flanders (FWO research project 1SHEN24N), by the BNP Paribas Fortis Chair in Fraud Analytics, and by the Flemish Government under the ``Onderzoeksprogramma Artificiële Intelligentie (AI) Vlaanderen'' programme.The resources and services used in this work were provided by the VSC (Flemish Supercomputer Center), funded by the Research Foundation - Flanders (FWO) and the Flemish Government.

\bibliographystyle{plainnat} 
\bibliography{bibliography.bib}

\end{document}